\documentclass[letterpaper]{article} 
\usepackage{aaai2026}  
\usepackage{times}  
\usepackage{helvet}  
\usepackage{courier}  
\usepackage[hyphens]{url}  
\usepackage{graphicx} 
\usepackage{natbib}  
\usepackage{caption} 
\usepackage{algorithm}
\usepackage{algorithmic}
\usepackage{amsmath, amssymb, amsthm}
\usepackage{newfloat}
\usepackage{listings}
\DeclareCaptionStyle{ruled}{labelfont=normalfont,labelsep=colon,strut=off} 
\floatstyle{ruled}
\newfloat{listing}{tb}{lst}{}
\floatname{listing}{Listing}
\usepackage{amsmath}
\usepackage{xcolor}
\usepackage{subcaption}
\usepackage{multirow}
\usepackage{booktabs}
\nocopyright

\title{Parametrized Multi-Agent Routing via Deep Attention Models}
\author{
    Salar Basiri\textsuperscript{\rm 1}\thanks{Corresponding author: sbasiri2@illinois.edu},
    Dhananjay Tiwari\textsuperscript{\rm 1},
    Srinivasa M. Salapaka\textsuperscript{\rm 1}
}
\affiliations{
    \textsuperscript{\rm 1}Coordinated Science Laboratory (CSL), University of Illinois Urbana-Champaign\\
}

\usepackage{bibentry}

\newcommand{\mc}[1]{\mathcal{#1}} 
\newcommand{\mb}[1]{\mathbb{#1}} 
\newcommand{\cb}[1]{\left\{#1\right\}} 
\newcommand{\rcb}[1]{\left[#1\right]} 
\newcommand{\rb}[1]{\left(#1\right)} 

\begin{document}

\maketitle

\begin{abstract}
We propose a scalable deep learning framework for parametrized sequential decision-making (ParaSDM), where multiple agents jointly optimize discrete action policies and shared continuous parameters. A key subclass of this setting arises in Facility-Location and Path Optimization (FLPO), where {\em multi-agent systems} must {\em simultaneously} determine optimal routes and facility locations, aiming to minimize the cumulative transportation cost within the network. FLPO problems are NP-hard due to their mixed discrete-continuous structure and highly non-convex objective.
To address this, we integrate the Maximum Entropy Principle (MEP) with a neural policy model called the Shortest Path Network (SPN)—a permutation-invariant encoder–decoder that approximates the MEP solution while enabling efficient gradient-based optimization over shared parameters.
The SPN achieves up to 100$\times$ speedup in policy inference and gradient computation compared to MEP baselines, with an average optimality gap of approximately 6\% across a wide range of problem sizes. Our FLPO approach yields over 10$\times$ lower cost than metaheuristic baselines while running significantly faster, and matches Gurobi’s optimal cost with annealing at a 1500$\times$ speedup—establishing a new state of the art for ParaSDM problems. These results highlight the power of structured deep models for solving large-scale mixed-integer optimization tasks.
\end{abstract}

\begin{links}
    \link{Code}{https://github.com/salar96/LearningFLPO}
\end{links}

\section{Introduction}
Combinatorial optimization problems (COPs) are central to numerous real-world applications, including logistics, scheduling, network design, and resource allocation. In recent years, there has been growing interest in applying machine learning (ML) techniques to address these problems more efficiently and at scale. However, COPs pose significant challenges for ML due to their discrete structure, combinatorial decision spaces, and complexity. Key limitations include poor scalability and generalization to larger or unseen instances, difficulty in generating labeled data, and lack of performance guarantees. Additionally, encoding constraints, ensuring interpretability, and integrating ML with traditional solvers add further complexity.

Despite these obstacles, ML methods—especially deep learning—have shown promise in automating heuristic design for COPs, which was traditionally reliant on manual, domain-specific expertise. Deep learning offers an end-to-end, data-driven alternative that has been successfully applied to problems like the Traveling Salesperson Problem (TSP), Vehicle Routing Problem (VRP), Knapsack Problem, and Set Covering~\cite{LeCun2015-al, wang2024solving, vinyals2017pointernetworks, bello2017neuralcombinatorialoptimizationreinforcement, kool2019attentionlearnsolverouting, bengio2020machinelearningcombinatorialoptimization}.

\emph{Neural combinatorial optimization (NCO)} provides a unified framework by modeling COPs as sequential decision-making (SDM) processes. Here, policies—defined as probability distributions over actions conditioned on the current state—are learned using deep neural networks, often trained via reinforcement learning. While this approach enables flexible, end-to-end modeling, it faces challenges such as sparse reward signals, slow convergence, and limited transferability across problem variants.

A relatively underexplored subclass of COPs is \emph{parameterized sequential decision-making (ParaSDM)}—a mixed discrete-continuous formulation that arises in applications such as supply chain design, transportation, robotics, and sensor networks. In ParaSDM, $N$ agents make sequential decisions over $K$ time steps, where the states, actions, and cost functions depend on a shared set of continuous parameters $\mathcal{Y} \subset \mathbb{R}^d$. At each step $k$, agent $i$ selects an action $a_k^i \in \mathcal{A}_\mathcal{Y}$, where the action space is itself parameterized by $\mathcal{Y}$. The optimization thus involves simultaneously solving for both the continuous parameters $\mathcal{Y}$ and the discrete policies ${a_k^j(\mathcal{Y})}$, resulting in a tightly coupled and highly non-convex optimization landscape.

\begin{figure*}[t] 
    \centering
    \begin{subfigure}[b]{0.24\textwidth}
        \includegraphics[width=\textwidth]{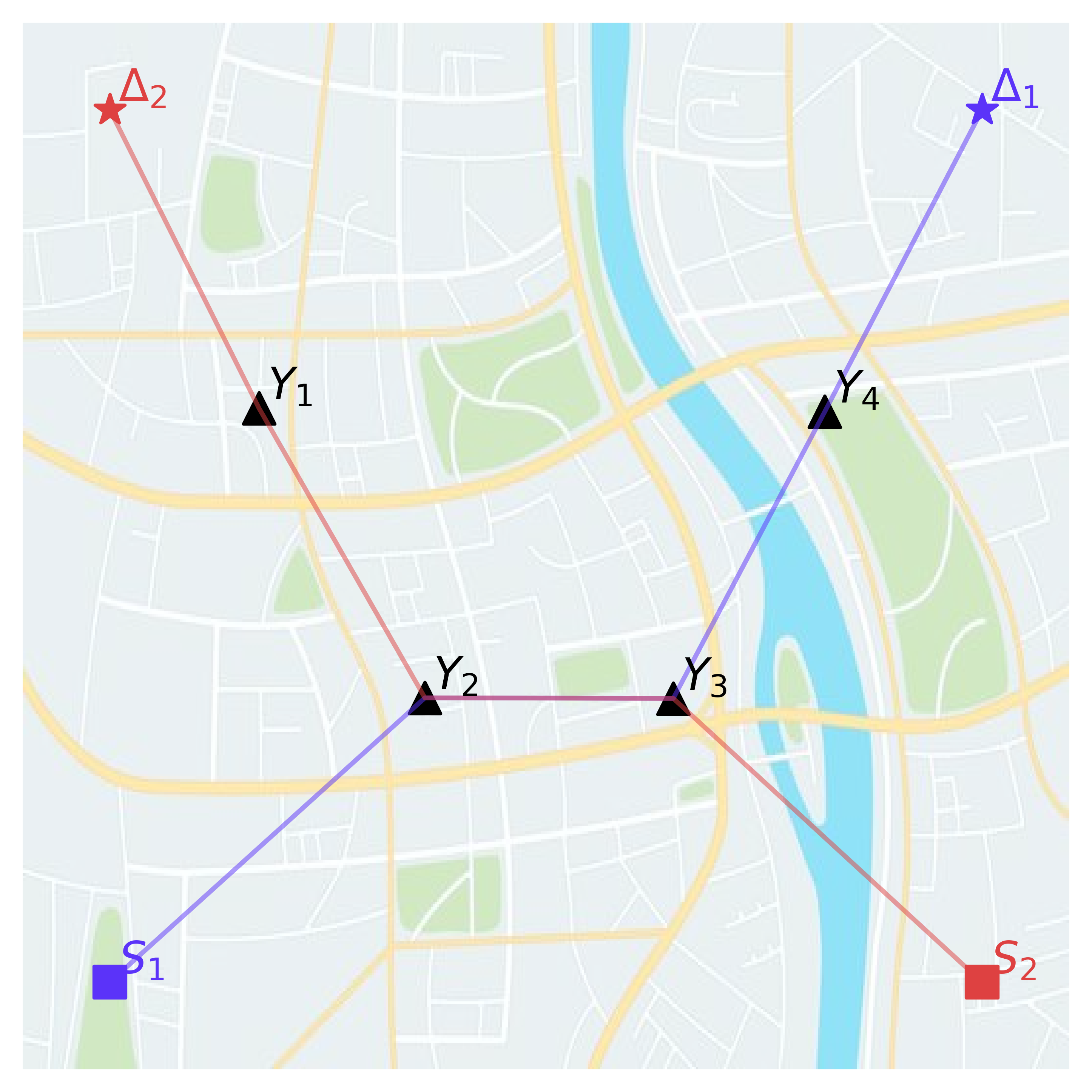}
    
        \label{fig:sub1}
    \end{subfigure}
    \hfill
    \begin{subfigure}[b]{0.24\textwidth}
        \includegraphics[width=\textwidth]{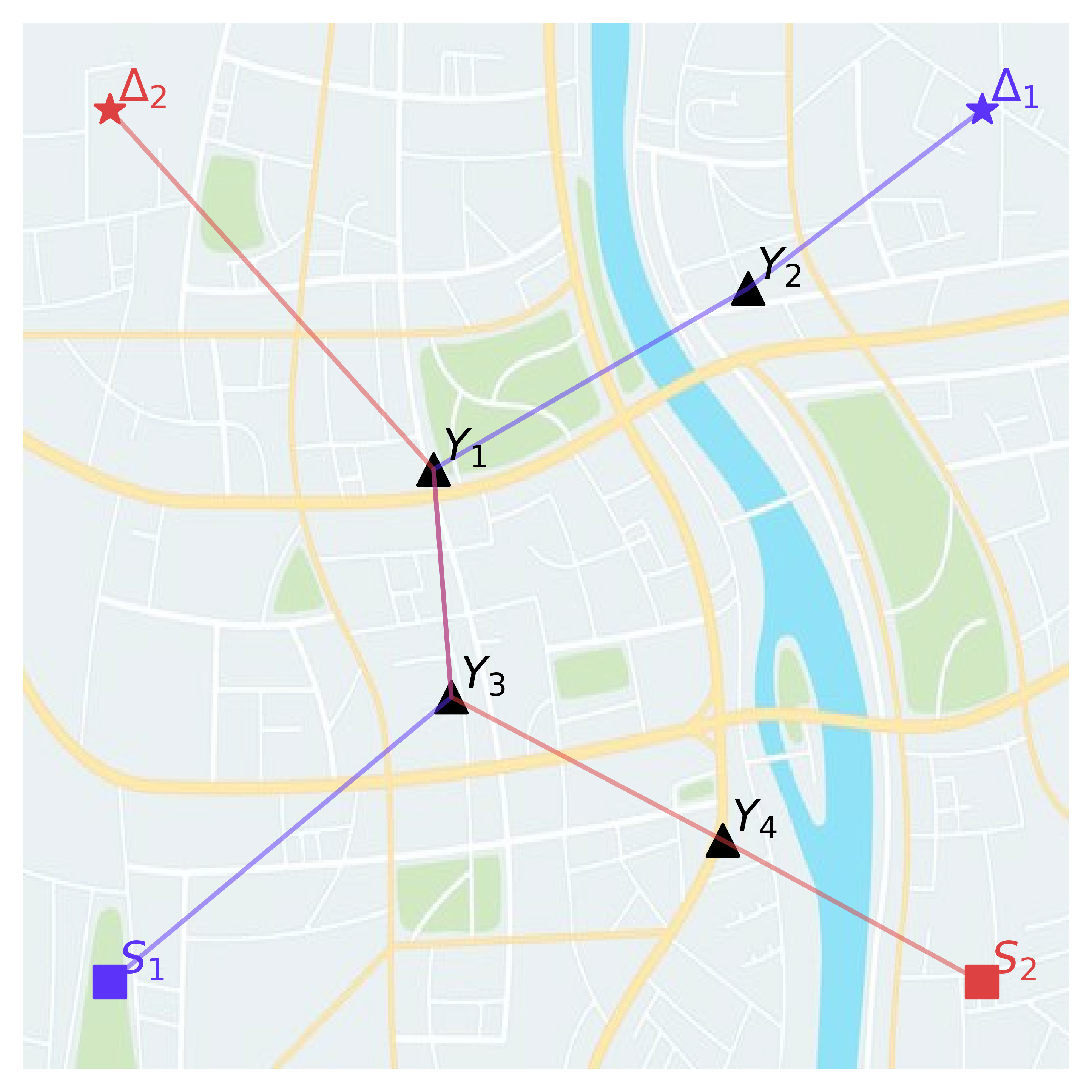}
        
        \label{fig:sub2}
    \end{subfigure}
    \hfill
    \begin{subfigure}[b]{0.24\textwidth}
        \includegraphics[width=\textwidth]{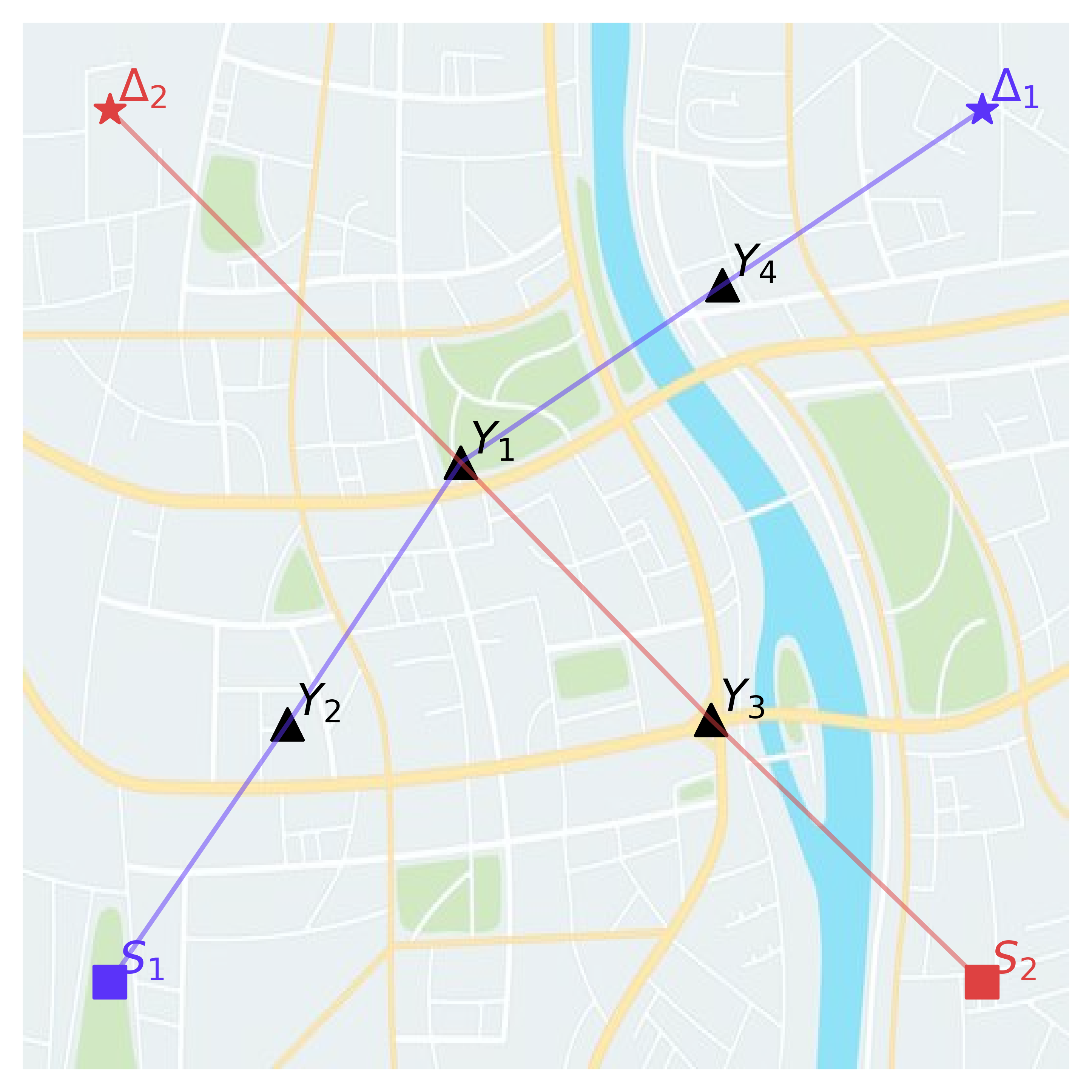}
        
        \label{fig:sub3}
    \end{subfigure}
    \hfill
    \begin{subfigure}[b]{0.24\textwidth}
        \includegraphics[width=\textwidth]{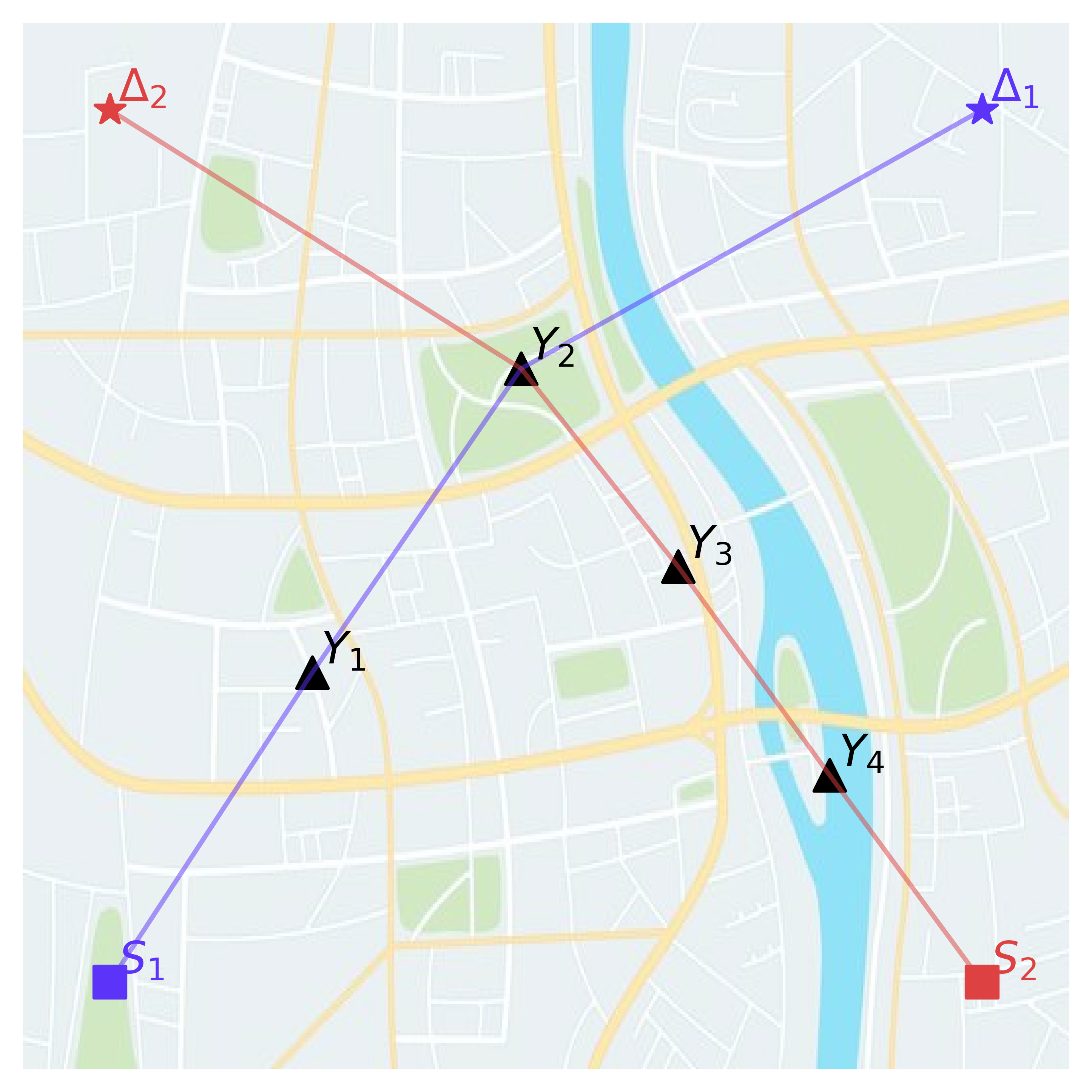}
        
        \label{fig:sub4}
    \end{subfigure}
    \caption{Multiple local minima shown for a small-scale FLPO problem ($S_i,\Delta_i$ represent start and end locations respectively). }
    \label{fig:FLPO Toy Problems}
\end{figure*}

A key subclass of ParaSDM is the \emph{facility-location and path optimization (FLPO)} problem, where the objective is to compute optimal travel paths for $N$ agents (e.g., drones) through a network of $M$ facilities (e.g., charging stations), while simultaneously determining the facility locations. Each agent must reach a designated destination, and the goal is to minimize total travel cost by jointly optimizing the discrete paths ${a_k^i}$ and the continuous node locations $\mathcal{Y}$. While path optimization alone can be modeled as a dynamic program or Markov decision process (MDP), jointly optimizing both paths and node locations creates a complex mixed-integer, non-convex problem. This problem is NP-hard, and even for small instances (e.g., $N=2$, $M=4$), the solution space contains numerous local optima, as illustrated in Figure~\ref{fig:FLPO Toy Problems}. In this setting, standard MDPs can be seen as special cases of ParaSDM where the continuous parameters (e.g., facility locations) are fixed.

Most NCO approaches in the ML literature have been applied to problems with fixed state and action spaces, where decisions are made over discrete choices (e.g., in TSP and VRP), and not conditioned on a separate set of continuous parameters $\mc Y$ that must be jointly optimized. Moreover, while effective, many of these models function as black-box predictors, often disregarding the underlying algebraic structure of the problem due to the intractability of analytic solutions. This leads to purely data-driven input-output mappings with limited interpretability and less integration of domain knowledge. 

In \cite{AmberFLPO,AmberparaMDP}, a framework based on the maximum entropy principle (MEP) \cite{jaynes2003probability} is developed to address ParaSDM. MEP has proved successful in addressing a variety of COPs, including facility location problems,   image processing, vector quantization, and other resource allocation problems. 
MEP based solution comprise minimizing a \emph{Free Energy}  $\mathcal{F}_\beta(\mathcal{Y},P)$, a regularized cost function parameterized with an {\em annealing parameter} $\beta$. Here the decision variables $\mathcal{Y}$ and $P$ respectively represent the facility location parameters and the path or policy variables. The Free Energy function is derived from the underlying cost function of the ParaSDM problem by relaxing a set of binary policy variables with soft probability associations $P$. In fact, $\mathcal{F}_\beta$ is the same as the ParaSDM cost function when $\beta\rightarrow \infty$.

The MEP-based algorithm solves for the best $\rb{\mc{Y},P}$ at every annealing step $\beta_k$ using inherent Gibbs distribution structure for $P$ and standard gradient-based optimization schemes for $\mc Y$. The solution at $\beta_k$ is used as an initial guess for the succeeding step $\beta_{k+1}$. This algorithmic solution is insensitive to initialization, robust to noisy data, can easily incorporate various component-node dynamic, communication, and capacity constraints. The annealing process promotes exploration of the cost landscape in the initial iterates and appropriate exploitation as the parameter $\beta$ is increased. This approach has demonstrated excellent results when applied to many NP-hard problems \cite{baranwal2022unified}. However it does not scale well for  ParaSDMs with large $N$ and $M$, where at each annealing step $\beta_k$, there are $O\left(N \sum_{k=1}^M \binom{M}{k}k!\right)$ policy variables for each fixed configuration of node locations $\mathcal{Y}$. 

This paper presents a scalable deep learning framework for solving FLPO problems, grounded in the algebraic structure of MEP. Neural architectures for ParaSDM pose unique challenges, including varying problem sizes and the need to preserve MEP properties such as annealing-driven exploration and robustness. 
A core computational bottleneck is the gradient of Free Energy function, which aggregates the gradients of travel costs over all paths, weighted by the Gibbs distribution form of the policies $P$. To address this, we introduce the Shortest Path Network (SPN)—a permutation-invariant encoder-decoder deep model that approximates Free Energy gradients by exploiting the Gibbs distribution structure of policies. 
At high $\beta$, SPN emphasizes the gradients of few dominant paths (exploitation), while at low $\beta$, it is augmented with a uniform sampling approach to promote broader exploration. The encoder uses linear self-attention blocks, and our novel decoder design applies a lightweight gating mechanism to fuse current and terminal states before cross-attending to facility nodes.

We propose two solver variants: a fast, initialization-sensitive method using SPN alone, and a more robust, higher-cost approach that closely approximates the full MEP solution. Although classical algorithms like Dijkstra or Bellman–Ford can find a single shortest path efficiently, MEP-based inference requires access to the top‑$b$ paths—something classical methods handle poorly without repeated re-execution or complex variants. In contrast, SPN enables efficient top‑$b$ inference via sampling or beam search over its learned policy. Its encoder–decoder architecture achieves sub-quadratic runtime in practice, with worst-case complexity of $O(NM^2)$, and is highly parallelizable across agents and graph instances—making it ideal for large-scale, real-time FLPO deployments.

Our Deep FLPO solution achieves significant scalability over the original MEP-based formulation, offering up to $100\times$ policy inference and differentiation without compromising solution quality. The SPN generalizes across varying network sizes (10-300 nodes) with an average optimality gap of 6\%. In benchmarks, Deep FLPO achieved over 10$\times$ lower cost than metaheuristic algorithms, while running significantly faster. Compared to Gurobi baseline, it reached within 2\% of optimality at high $\beta$, and matched Gurobi’s cost with annealing—at roughly 1500$\times$ speedup. These results demonstrate Deep FLPO’s efficiency and establish it as a strong state-of-the-art method for paraSDM problems.

\section{Background}
Much of the machine learning literature in this domain focuses on classical combinatorial problems such as the TSP and VRP, which are closely related to our task of learning shortest paths between designated START and END nodes. ML-based approaches aim to train deep models that can directly predict high-quality solutions for new instances without requiring optimization at inference time. However, this paradigm faces several fundamental challenges: the model must generalize across varying network sizes and topologies; generating high-quality supervision is computationally expensive; and the highly combinatorial, multimodal nature of the solution space makes it difficult for neural models to capture and navigate effectively.

 Several key works have advanced the use of neural models for combinatorial optimization by tackling these core challenges. The Pointer Network (PN) \cite{vinyals2017pointernetworks} introduced attention-based mechanisms to handle variable-sized inputs and output permutations in a supervised setting. This was later extended by \cite{bello2017neuralcombinatorialoptimizationreinforcement}, who used an Actor-Critic reinforcement learning framework with LSTMs to eliminate the need for expensive supervision. Building on this, \cite{nazari2018reinforcementlearningsolvingvehicle} simplified the architecture for VRP by removing recurrence in favor of element-wise embeddings, making the model more scalable and easier to train.

A major breakthrough came with fully attention-based models, particularly the Attention Model (AM) of \cite{kool2019attentionlearnsolverouting}, which replaced recurrent components with a Transformer-style encoder–decoder. This shift enabled greater parallelism and improved scalability, while representing the solution as a stochastic policy over sequences to better explore the multimodal solution space. {\color{black}Further inference speed-up was made in \cite{joshi2019efficient} using Graph Neural Network (GNN) representations of TSP and parallelized non-autoregressive decoding.}


Several works have extended attention-based models to better handle the permutation-invariant nature of problems like TSP. Deep Sets \cite{zaheer2018deepsets} embed elements independently with permutation-invariant pooling, while Set Transformers \cite{lee2019set} reduce attention complexity using learnable inducing points, though their fixed-size decoder limits flexibility. \cite{bresson2021transformernetworktravelingsalesman} modified Transformers with batch normalization and a learnable start token, improving TSP performance. More recently, \cite{jung2024lightweightcnntransformermodellearning} introduced a hybrid CNN-transformer that aggregates local node features and restricts decoder attention to the last few visited nodes, offering significant speedups with strong solution quality.

Another major challenge addressed is training NCO agents at scale: supervised learning needs costly labels, while reinforcement learning (RL) struggles with sparse rewards. Some approaches aim to generalize from small instances \cite{luo2024LEHD,drakulic2023bqncobisimulationquotientingefficient}, decompose large problems into smaller ones \cite{hou2023generalize,ye2024gloplearningglobalpartition}, {\color{black}or use a hybrid GNN-local search approach \cite{hudson2021graph} to generalize to larger instances without requiring fully supervised training.} A recent alternative, self-improved learning (SIL) \cite{luo2024selfimprovedlearningscalableneural}, avoids labels and scales up efficiently using linear-time attention. RL-based methods also enhance combinatorial solvers by guiding local search via neural policies \cite{wu2021learning,costa2020learning,chen2019learning,hottung2020neural,lu2019learning}, with policy gradient techniques preferred for their stability and adaptability to large or continuous action spaces.

While there is extensive work on learning policies to optimize agent actions over discrete sets given a fixed problem instance, far less attention has been given to ParaSDM problems—where the state and action spaces are coupled with a set of continuous parameters that must be co-optimized. Various approaches that independently determine the parameters $\mc Y$ and subsequently find the optimal policies often yield suboptimal solutions that are sensitive to initialization and prone to bias.

In our prior work, we addressed such ParaSDM problems using MEP, extending it to both finite \cite{AmberFLPO} and infinite horizons \cite{AmberparaMDP}, with applications in UAV-UGV coordination, small cell placement, and joint routing-scheduling \cite{Scitech1,scitech2}. These approaches substantially improved robustness and solution quality across a variety of constrained settings. However, in the finite-horizon ParaSDM framework \cite{AmberFLPO}, agent policies and cost gradients are computed via dynamic programming, which is polynomial in graph size $M$ and must be recomputed at every parameter update $\mathcal{Y}$. This makes scaling to even moderately large instances computationally expensive.

This paper bridges the gap between scalable ML approaches—which often disregard structural priors—and the MEP-based ParaSDM framework—which encodes rich structure yet faces scalability barriers. Our key contribution is to fuse an NCO architecture with the algebraic structure of MEP, enabling joint learning of agent policies and shared network parameters in a scalable and structure-aware manner.

\section{Problem Formulation}
\label{sec:problem_formulation}

We consider a scenario of $N$ agents along with $M$ spatial nodes in a domain $\mc D \subset \mb R^d$.
Each agent $v_i, 1 \leq i \leq N$ travels from $s_i \in \mc D$ (START) to its destination $\Delta_i \in \mc D$ (END). Each agent is assigned a weight $\rho_i \in [0 , 1]$ according to its relative importance. The agents are routed through the set of spatial nodes (facilities) $f_j, \ 1 \leq j \leq M$ located at $y_j \in \mc D$. We define $\mc Y \in \mc D^M$ as the concatenated vector of all the node locations $y_j$. The total travel cost incurred by an agent along a path is taken as the cumulative sum of the square Euclidean distance between the consecutive nodes.

\begin{figure}[t]
    \centering
    \includegraphics[width=0.9\linewidth]{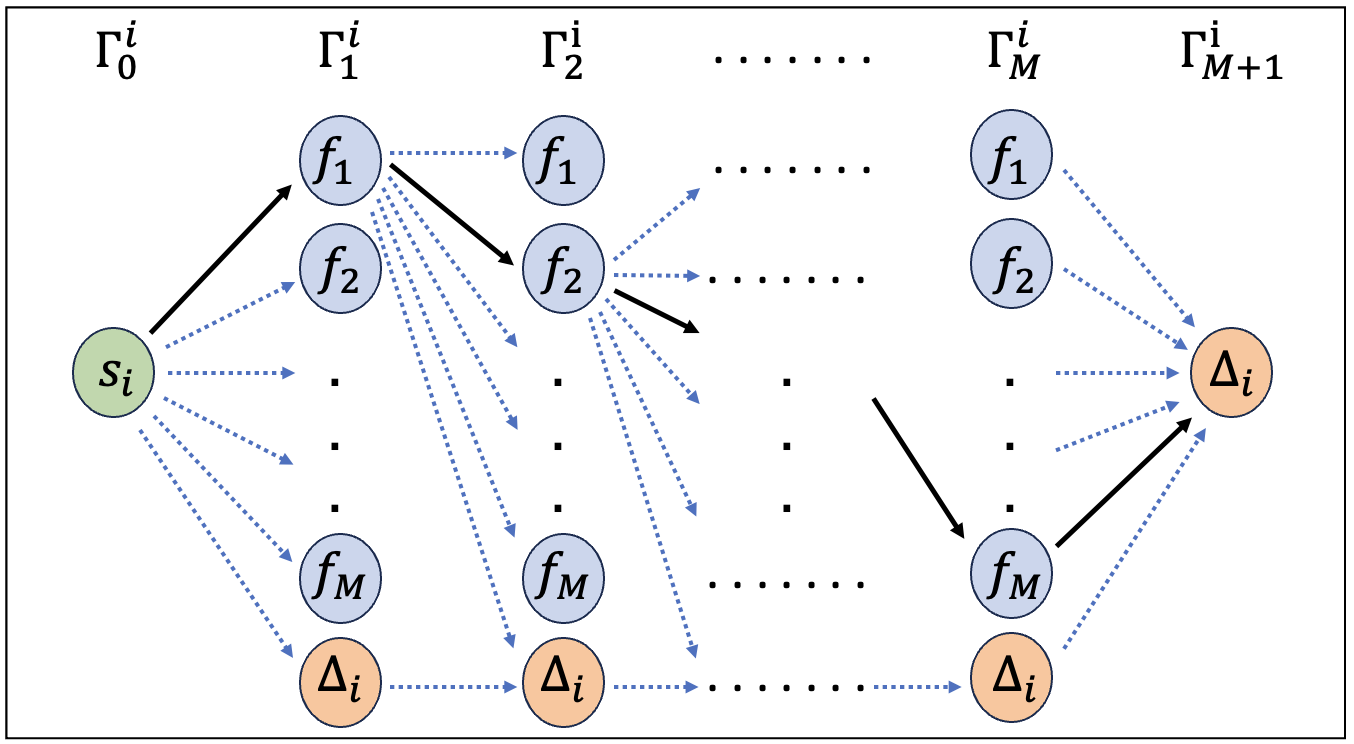}
    \caption{A finite-horizon stagewise architecture for agent transitions between start and goal locations through nodes.}
    \label{fig:flpostructure}
\end{figure}

The transition of each agent from $s_i$ to $\Delta_i$ via nodes at $\mc Y$ is modeled in a finite-horizon FLPO architecture (as shown in Figure \ref{fig:flpostructure}). 
The architecture consists of $M+2$ stages $\Gamma_{k}^i, 0 \leq k \leq M+1$, such that $\Gamma_0^i = \cb{s_i}, \ \Gamma_k^i = \cup_{j=1}^M \cb{f_j} \cup \cb{\Delta_i}, 1 \leq k \leq M$ and $\Gamma_{M+1}^i = \cb{\Delta_i}$
The path of each agent is characterized by $\gamma \in \mc G^i$, where $\mc G^i$ denotes the space of all possible paths between $s_i$ and $\Delta_i$,
\begin{align*}
    \mc G^i := \cb{\gamma :=\rb{\gamma_1,\gamma_2,\dots,\gamma_M} \mid \gamma_k \in \Gamma_k^i, 1 \leq k \leq M}.
\end{align*}

The aim is to minimize the following transportation cost, 
\begin{align}
    \min_{\substack{\mc Y, \eta^i\rb{\gamma}}} \quad & D := \sum_{i=1}^N \rho_i \sum_{\gamma \in \mc G^i} \eta^i\rb{\gamma} d^i\rb{\gamma}, \label{eq:hard_objective} \\
    \textrm{ subject to}\quad & \eta^i\rb{\gamma} \in \cb{0,1}, \forall \gamma, \ \sum_{\gamma \in \mc G^i} \eta^i\rb{\gamma} = 1, \forall i \label{eq:hard_path_cons}
\end{align}
where for each agent $v_i$, $d^i:\mc G^i \rightarrow [0,\infty)$ denotes the transportation cost along path $\gamma$, and $\eta^i:\mc G^i \rightarrow \cb{0,1},$ represents the decision to choose path $\gamma$, with $\eta^i\rb{\gamma} = 1$ if $\gamma$ is chosen, $\eta^i\rb{\gamma} = 0$ otherwise. The path that minimizes $d^i\rb{\gamma}$ depends on node locations $\mc Y$. 

For a total of $N$ agents, the optimal paths of agents are selected from a set of size $O(N\sum_{k=1}^M \binom{M}{k}k!)$ \emph{for each instance of node locations} $\mc Y$. 
This results in a mixed-integer optimization problem that is NP-hard and highly non-convex. 

The approaches in \cite{AmberFLPO, AmberparaMDP} relax the hard associations $\eta^i$ by introducing soft probabilistic assignments $p^i : \mc G^i \rightarrow [0,1]$. 
They reformulate the original problem \eqref{eq:hard_objective}–\eqref{eq:hard_path_cons} as the minimization of a regularized cost $F_\beta$ (with parameter $\beta$) given by,
\begin{align}
    F_{\beta} := \sum_{i=1}^N \rho_i \sum_{\gamma \in \mc G^i}  p^i\rb{\gamma} \rcb{ d^i\rb{\gamma} + \frac{1}{\beta} \ln p^i\rb{\gamma}} \label{eq:free_energy}
\end{align}
jointly over the node locations $\mc Y$ and soft path assignments ${p^i}$. 
The cost $F_\beta$ consists of the relaxed objective ($D$) and the total Shannon entropy ($H$) of the distributions $p^i$, weighted by parameter $1/\beta$. Thus, $F_\beta$ captures a tradeoff between approximation accuracy and sensitivity to initializations. 
At lower $\beta$, $F_\beta$ is convex due to dominant $H$ term and a global minimum $\rb{\mc Y_0, p^i_0}$ of $F_\beta$ is located. At this global minimum, all paths are treated with equal priority, and the node distributions are identical across all agents. As $\beta$ is increased (moving from exploration to exploitation), $F_\beta$ is minimized iteratively by using the solutions $(\mc Y_\beta, p^{i}_\beta)$ from previous $\beta-$iterations as initial condition. As $\beta \rightarrow \infty$, the entropy term vanishes, the soft associations $p^i$ converge to hard assignments $\eta^i$, and the regularized cost $F_\beta$ reduces to the original objective $D$ in \eqref{eq:hard_objective}, thereby recovering a solution to the original problem.
 

A key advantage of this approach is that the joint optimization over $\rb{\mc Y, {p^i}}$ reduces to an optimization solely over the node locations $\mc Y$, as the first-order optimality condition for $p^i$ admits a closed-form solution, given by the Gibbs distribution:
\begin{align}
    p^i_\beta(\gamma) = \frac{\exp\rcb{-\beta d^i(\gamma)}}{\sum_{\gamma' \in \mc G^i} \exp\rcb{-\beta d^i(\gamma')}}, \quad \forall i. \label{eq:gibbs_p}
\end{align}
Consequently, at each $\beta$, an optimal set of node locations $\mc Y_\beta$ is obtained by implementing a gradient-based update scheme $Q$ of the form,
\begin{align}
    \mc Y^{t+1} & = Q\rb{\mc Y^t, \nabla_{\mc Y^t}F_\beta}, \quad  t=0,1,2,\dots \label{eq:y_update_scheme} \\
    \textrm{where }
    \nabla_{\mc Y^t}F_\beta & := \sum_{i=1}^N \rho_i \sum_{\gamma \in \mc G^i} p^i_\beta\rb{\gamma} \nabla_{\mc Y^t}{d}^i\rb{\gamma},\label{eq:gradF_y_1}
\end{align}
results from differentiating Eq~\eqref{eq:free_energy} with respect $\mc Y^t$ for each update step $t$.
Since the gradient computation cost per step in Eq~\eqref{eq:gradF_y_1} scales as $O\rb{NM^2\sum_{k=1}^M \binom{M}{k}k!}$, a Markov property assumption is used to dissociate the assignments over the paths into a set of stage-wise associations, i.e., $p^i\rb{\gamma} = \prod_{k=0}^{M-1} p^i\rb{\gamma_{k+1}|\gamma_k}, \forall k$. This yields a set of (parametrized) dynamic programming (DP) iterations across the stages $\Gamma_k^i$, at each $\mc Y^t$ to evaluate gradient in Eq~\eqref{eq:gradF_y_1}, reducing the worst-case scaling to $O(NM^4)$ (see Appendices \ref{APP:value_funcs_free_energy}-\ref{APP:complexity_grad_stagewise} for details). Although the approach scales the problem for multi-agent settings and circumvents the initialization biases compared to the traditional solvers (as demonstrated in \cite{Scitech1}, \cite{scitech2}), the \textit{repeated} dynamic programming (DP) to perform updates via Eqs~\eqref{eq:y_update_scheme}-\eqref{eq:gradF_y_1} becomes increasingly expensive even for moderately large values of $M$.
To address the scalability challenges of FLPO—stemming from both the dynamic programming nature of policy computation and, more critically, the repeated computation of cost gradients with respect to parameters—we propose a neural combinatorial optimization model called the Shortest Path Network (SPN). 
Given agent start locations ${s_i}$, destinations ${\Delta_i}$, and network parameters ${\mc Y^t}$, SPN approximates the stepwise policy $p_k^i(\gamma_{k+1}|\gamma_k) \ \forall k$ with worst-case complexity $O(NM^2)$ and significant hardware-level parallelism across agents. 
SPN then enables efficient sampling of actions within a fully differentiable PyTorch framework, where the cost function and network parameters ${\mc Y^t}$ are embedded in the computation graph. 
This setup allows immediate and parallel gradient computation across all agents, bypassing the need for traditional solvers and external differentiation.

SPN is trained to mimic the Gibbs distribution (\ref{eq:gibbs_p}) in the high-$\beta$ regime, recovering deterministic shortest paths with near-zero entropy. While this allows for fast, purely neural FLPO inference at high $\beta$, such solutions are sensitive to initialization and may not reflect the annealed MEP behavior required for optimality. Crucially, our deep learning model also enables efficient extraction of a diverse set of top‑$b$ shortest paths for each agent in a single forward pass via sampling or beam search. This capability is central to the mixture sampling scheme introduced in the next section, which approximates the annealing behavior of FLPO, with improved robustness and solution quality at the cost of additional computation.


\subsection{Empirical Cost Distribution and Gradient Estimation}

To mimic the annealing behavior of FLPO solution introduced earlier, we employ a mixture sampling approach to implement the update scheme \eqref{eq:y_update_scheme}. 
At given $\beta$ and instance $\mc Y^t$ we estimate the gradients $\nabla_{\mc Y^t}F_\beta$ by sampling a set of $L$ paths $\mc G^i_L$ between $s_i$ and $\Delta_i$ for each agent $v_i$, 
$$\mc G^i_L = \cb{\gamma^1, \dots, \gamma^b, \gamma^{b+1},\dots,\gamma^{L}}.$$
The first $b$ paths $\cb{\gamma^1,\dots,\gamma^b}$ are obtained by using beam search or direct sampling the path nodes over the learned SPN policy, $\pi_\theta$, 
where $\theta$ denotes the SPN parameters (weights and biases).
The rest of the paths are sampled via a stage-wise uniform distribution, $\gamma_{k+1}^q \sim \textrm{unif}\rb{\Gamma_{k+1}^i|\gamma_k}, \forall k, b+1 \leq q  \leq L$. 

An estimate of the gradient in Eq~\eqref{eq:gradF_y_1} is computed as,
\begin{align*}
    \widehat{\nabla_{\mc Y^t} F_\beta} = \sum_{i=1}^N \rho_i \sum_{q=1}^{L} \hat p^i_\beta\rb{\gamma^q} \nabla_{\mc Y^t}d^i\rb{\gamma^q}, && \forall j, \forall r.
\end{align*}
where $\hat p^i_{\beta}\rb{\gamma^q}$ is a Gibbs distribution over the samples $\mc G^i_L$,
\begin{align*}
    \hat p^i_{\beta}\rb{\gamma^q} = \frac{\exp{\rcb{-\beta d^i\rb{\gamma^q}}}}{\sum_{q=1}^{L}\exp{\rcb{-\beta d^i\rb{\gamma^q}}}}, && \forall q, \forall i.
\end{align*}
The gradients $\nabla_{\mc Y^t} d^i(\gamma^q)$ are obtained through backpropagation, leveraging PyTorch’s automatic differentiation and computation graph.

A key aspect of this approach is estimating the first $b$ shortest paths accurately using SPN as the gradients $\nabla_{\mc Y^t}F_\beta$ are significantly influenced by shorter paths as $\beta$ is increased. This is because the Gibbs distribution assigns a higher probability to a shorter path i.e.,
\begin{align*}
    p^i_\beta\rb{\gamma_1} > p^i_\beta\rb{\gamma_2}, \textrm{ if } d^i\rb{\gamma_1} < d^i\rb{\gamma_2}, && \forall \gamma_1, \gamma_2, \forall i.
\end{align*}
As $\beta$ increases, the probability shifts to higher values towards lower costs and lower values towards higher costs, gradually assigning a probability of $1$ to the shortest path and $0$ to all the other paths, as $\beta \rightarrow \infty$.

Further, uniform sampling is necessary in order to have an efficient exploration and influence the gradients by as many facilities as possible in the initial phases of the $\beta-$iterations. This is necessary because as $\beta \rightarrow 0$ the Gibbs distribution prioritizes all the paths equally, i.e., $p^i\rb{\gamma} \rightarrow 1/|\mc G^i|, \forall \gamma, \forall i.$ Thus, uniform sampling enables the updates via Eq~\eqref{eq:y_update_scheme} avoiding many poor local minima at lower $\beta$ values.

In the following section, we present the deep learning approach to build SPN, the model architecture we have used and the training approach.

\section{Deep SPN Architecture} \label{Sec: SPN}
Inspired by Pointer Networks and Transformer-based models in neural combinatorial optimization, we adopt a simple encoder-decoder architecture to compute the action distribution $\pi_\theta^i(\gamma_{k+1} \mid \gamma_k)$ at each decision step $k$. \textcolor{black}{This determines the probability of an entire path $\gamma \in \mc G^i$ given $s_i$ and $\Delta_i,$ such that, $\pi^i_\theta\rb{\gamma} = \prod_{k=0}^{M-1} \pi^i_\theta\rb{\gamma_{k+1}|\gamma_k}, \forall i$. The architecture enables parallel computation of policies for all  $\rb{s_i, \Delta_i}$ and any instance of facility locations $\mc Y^t$.}

The encoder processes the concatenated Start ($S \in \mathbb{R}^{N \times d}$) and Destination ($\Delta \in \mathbb{R}^{N \times d}$) locations of all agents, along with the $M$ spatial node coordinates at the current instance ($Y^t \in \mathbb{R}^{N \times M \times d}$) constructed by broadcasting $\mc Y^t$ across all agents. Let $X\rb{k} \in \mb R^{N\times d}$ denote the concatenated position of all the agents at decision step $k$. Following induced attention mechanisms as in \cite{lee2019set}, the encoder stacks $L_{\text{enc}}$ layers of attention blocks over the concatenated input \textcolor{black}{$(S, Y^t, \Delta) \in \mb R^{N\times (M+2) \times d}$}.

\begin{figure}
    \centering
    \includegraphics[width=0.9\linewidth]{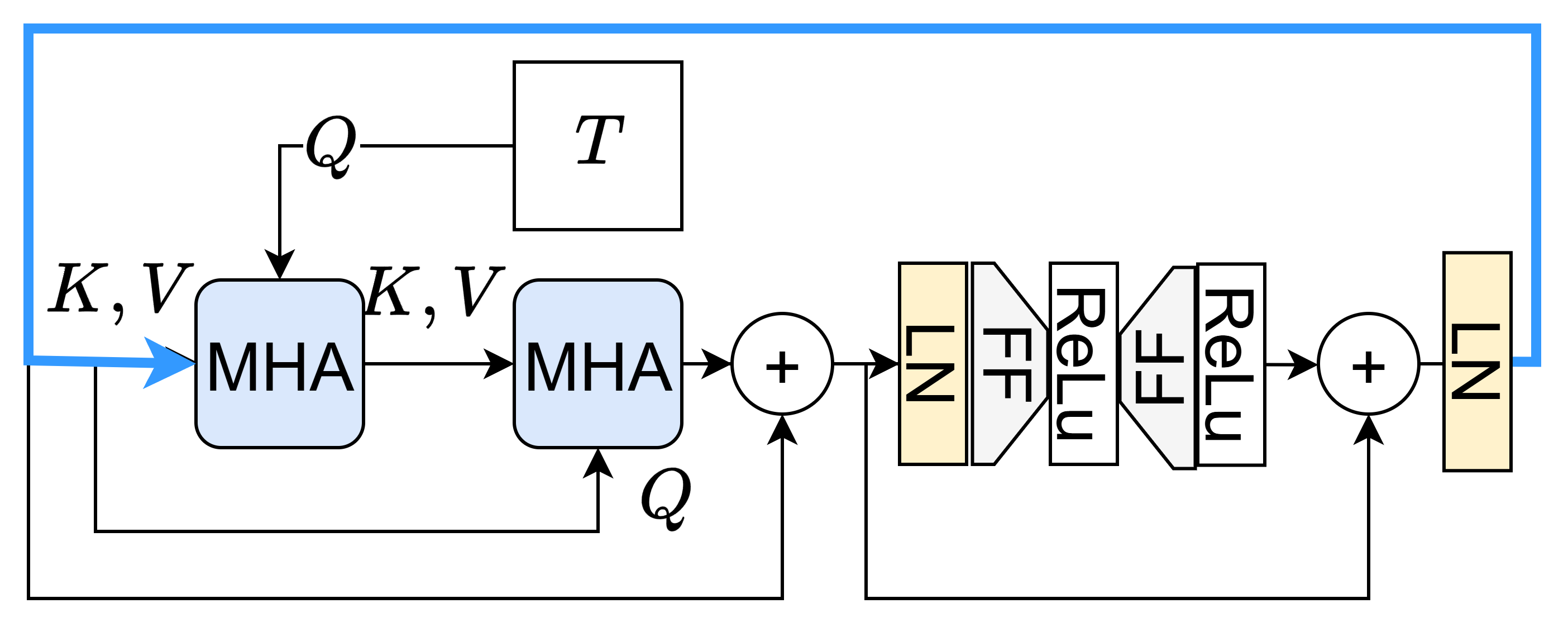}
    \caption{The Encoder Architecture.}
    \label{fig:Encoder}
\end{figure}

\textcolor{black}{A representation of the encoder architecture is shown in Figure \ref{fig:Encoder}}. Let $\text{MHA}(Q, K, V)$ denote the standard Multihead Attention for a query $Q$, key $K$ and value $V$, and $\text{LN}(\cdot)$ the LayerNorm operation. At layer $l$, the encoder output $E_l \in \mathbb{R}^{N \times (M + 2) \times d_h}$ is computed as:
\begin{align*}
    I_l &= \text{LN}\left(E_{l-1} + \text{MHA}(E_{l-1}, \hat{E}_{l-1}, \hat{E}_{l-1})\right), \\
    E_l &= \text{LN}\left(I_l + \mathcal{W}_{FF}(I_l)\right),
\end{align*}
where $\hat{E}_{l-1} = \text{MHA}(T, E_{l-1}, E_{l-1})$ uses a learned set of $M^*$ vectors $T \in \mathbb{R}^{M^* \times d_h}$. Here $M^*$ is a fixed and typically small hyperparameter. The initial embedding $E_0$ is formed by linearly projecting the concatenated inputs:
\[
    E_0 = \mathcal{W}_\text{IN}\left(\left[S, Y^t, \Delta\right]\right) \in \mathbb{R}^{N \times (M+2) \times d_h},
\]
where $\mathcal{W}_\text{IN}: \mathbb{R}^d \to \mathbb{R}^{d_h}$ is a linear layer with ReLU activation. The feedforward block $\mathcal{W}_{FF}: \mathbb{R}^{d_h} \to \mathbb{R}^{d_h}$ expands to $d_{FF} = 2d_h$ and contracts back to $d_h$, with ReLU activations in both transformations. Unlike the standard multihead self-attention which has quadratic time complexity $\mathcal{O}(NM^2)$, this architecture for encoder has $\mathcal{O}(NM\times M^*)$ complexity which is linear with respect to $M$.

To generate a goal-aware action distribution for selecting the next node, the decoder explicitly conditions on the current agent positions $X\rb{k} \in \mathbb{R}^{N \times d}$ and the destination $\Delta \in \mathbb{R}^{N \times d}$. Let $h_X, h_\Delta \in \mathbb{R}^{N \times d_h}$ denote the embeddings of the current and destination locations, respectively. We first construct a fused query representation $q$ via a gated interpolation:
\begin{align*}
    \alpha &= \sigma\left(\left[h_X, h_\Delta\right]W \right) \in [0, 1]^{N \times 1}, \\
    q &= \alpha \odot h_X + (\mathbf{1}^{N \times1} - \alpha) \odot h_\Delta \ \in \mb R^{N\times d_h},
\end{align*}
where $\sigma(\cdot)$ denotes the element-wise sigmoid function, $\odot$ denotes the Hadamard product (broadcast $d_h$ times) and $W \in \mathbb{R}^{2d_h \times 1}$ is a learnable parameter vector. The query $q$ attends over the encoder output $E_{L_{\text{enc}}} \in \mathbb{R}^{N \times (M+2) \times d_h}$ to compute the policy distribution for all agents:
\begin{align*}
    \Pi_k =
    \mathrm{softmax}\left(E_{L_{\text{enc}}}  \cdot q/\sqrt{d_h} \right) \in [0, 1]^{N \times (M+2)},
\end{align*}
where the dot product is taken along the $d_h$ dimension and $\Pi_k$ represents concatenation of action distributions $\pi^i_\theta\rb{\cdot|\gamma_k}, \forall k$ across all agents.

\begin{figure}
    \centering
    \includegraphics[width=0.9\linewidth]{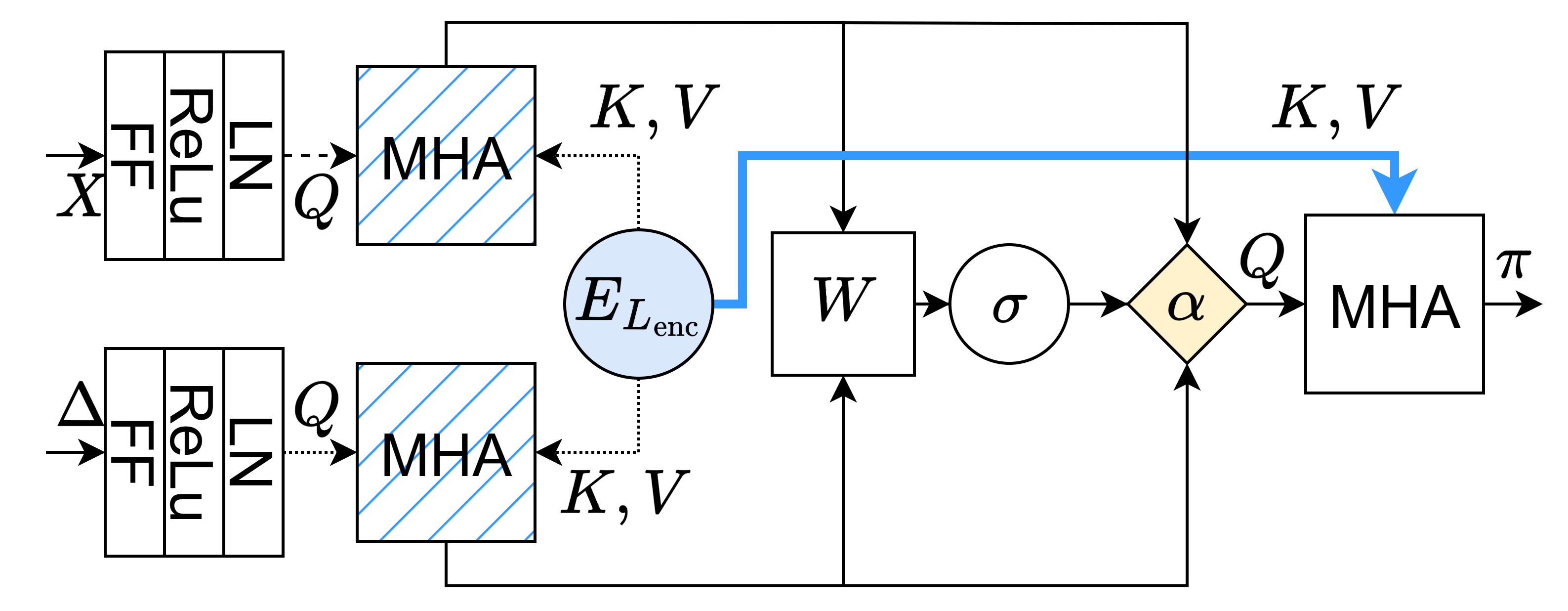}
    \caption{The decoder architecture. The DCAD model uses the hatched MHA blocks, while the DED model does not.}
    \label{fig:decoder}
\end{figure}

For ablation study, we consider two decoder variants for generating $h_X$ and $h_\Delta$:

\begin{itemize}
    \item \emph{Direct Embedding Decoder (DED)}: A minimal decoder that uses LayerNorm-transformed projections of the inputs:
    $
        z_S = \mathrm{LN}(\mathcal{W}_S(X\rb{k})), \ z_\Delta = \mathrm{LN}(\mathcal{W}_\Delta(\Delta)),
    $
    where $\mathcal{W}_S, \mathcal{W}_\Delta: \mathbb{R}^d \to \mathbb{R}^{d_h}$ are learnable linear layers with ReLU activations. We set $h_X = z_S$ and $h_\Delta = z_\Delta$.
    
    \item \emph{Dual Cross-Attention Decoder (DCAD)}: An alternative where the current and destination embeddings query the encoder output independently via separate cross-attentions:
    $
        h_X = \mathrm{MHA}(z_S, E_{L_{\text{enc}}}, E_{L_{\text{enc}}}), \
        h_\Delta = \mathrm{MHA}(z_\Delta, E_{L_{\text{enc}}}, E_{L_{\text{enc}}}).
    $ See Figure \ref{fig:decoder} for details.
\end{itemize}
The full path for each agent is generated auto-regressively: at each timestep, the decoder produces a probability distribution over candidate nodes, an action is realized (via sampling, greedy or beam search), and the agent transitions to the selected node. This process continues until the destination is reached.


\section{Training}
The SPN model is trained using a four-phase Curriculum Learning \cite{soviany2022curriculumlearningsurvey,BengioCurriculum} paradigm that progressively scales problem complexity while leveraging bootstrapped representations (details shown in table \ref{tab:training_scheme}). Phase 1 employs supervised learning on small graphs, establishing core policy priors. Phase 2 continues supervised training on larger graphs, initialized from Phase 1 to promote transfer. Phase 3 switches to RL via policy gradient (PG), comparing training from scratch versus leveraging Phase 2 pretraining. Then, Phase 4 scales training to 100-node graphs, further refining the pretrained policy. Finally, to improve generalization across varying graph sizes, we include a light Fine-Tuning (FT) stage on small graphs (10 nodes). This allows SPN to retain performance across both small and large scales, mitigating ``catastrophic forgetting" effects \cite{van2024continual} introduced during later RL phases.

\begin{table}[t]
\centering
\small  
\begin{tabular}{@{}l|l|l|l|l@{}}
\textbf{Phase} & \textbf{Mode} & \textbf{Pretrain} & \textbf{Epochs} & \textbf{M} \\ \hline
1 & Supervised & None & 10k & 10 \\
2 & Supervised & Phase 1 & 1.5k & 50 \\
3 & RL (PG) & None/Phase 2 & 50k & 50 \\
4 & RL (PG) & Phase 3 & 10k & 100 \\ \midrule
FT & RL (PG) & Phase 4 & 1k & \{10,100\}
\end{tabular}
\caption{Curriculum training scheme. Learning rate: $10^{-4}$, batch size: 256.}
\label{tab:training_scheme}
\end{table}

The decoder is trained to imitate the stagewise Gibbs policy defined in Eq.~\eqref{eq:gibbs_p}, which depends on an inverse temperature parameter $\beta$. Let $\theta$ denote the learnable parameters of the SPN. 
The supervised learning objective minimizes the expected KL divergence between the model policy $\pi^i_\theta(\gamma)$ and the target distribution $p_\beta^i(\gamma)$:
\begin{align} \label{eq:supervised cost}
    \min_\theta \ \mathbb{E}_{v_i \sim \mc P}\left[ \, \mathbb{E}_{\gamma \sim \mathcal{G}^i} \left[\mathbf{D}_{\text{KL}} \left(\pi^i_\theta(\gamma) \, \| \, p_\beta^i(\gamma)\right)\right] \right],
\end{align}
where the outer expectation is taken on varying agent instances $v_i = (s_i,\mc Y, \Delta_i)$, sampled uniformly from a probability density function $\mc P: \mc D^{M+2} \to \mathbb{R}.$  
The temperature $\beta$ controls the entropy of the ground-truth policy: $\beta \to 0$ induces uniform sampling, while $\beta \to \infty$ concentrates all probability mass on shortest paths. 
To facilitate optimization, we employ a KL bootstrapping strategy by gradually increasing $\beta$ during training (annealing), starting from low values where the random initialization of $\theta$ already yields high-entropy policies close to the target. Both expectations in Eq.~\eqref{eq:supervised cost} are estimated via Monte Carlo sampling. The effect of annealing is evaluated in the ablation study.

While the supervised phase encourages imitation of stochastic shortest paths, it may not capture global dependencies or longer-term optimality, especially when the graph size increases. As the model improves, especially on small or moderately-sized graphs, the supervised loss saturates — the KL divergence flattens out, providing little gradient signal to further refine the policy. This is especially evident when the model begins to approximate the optimal solution distribution and supervision no longer offers meaningful corrections. Thus, we refine the model using reinforcement learning.

\begin{figure}
    \centering
    \includegraphics[width=0.99\linewidth]{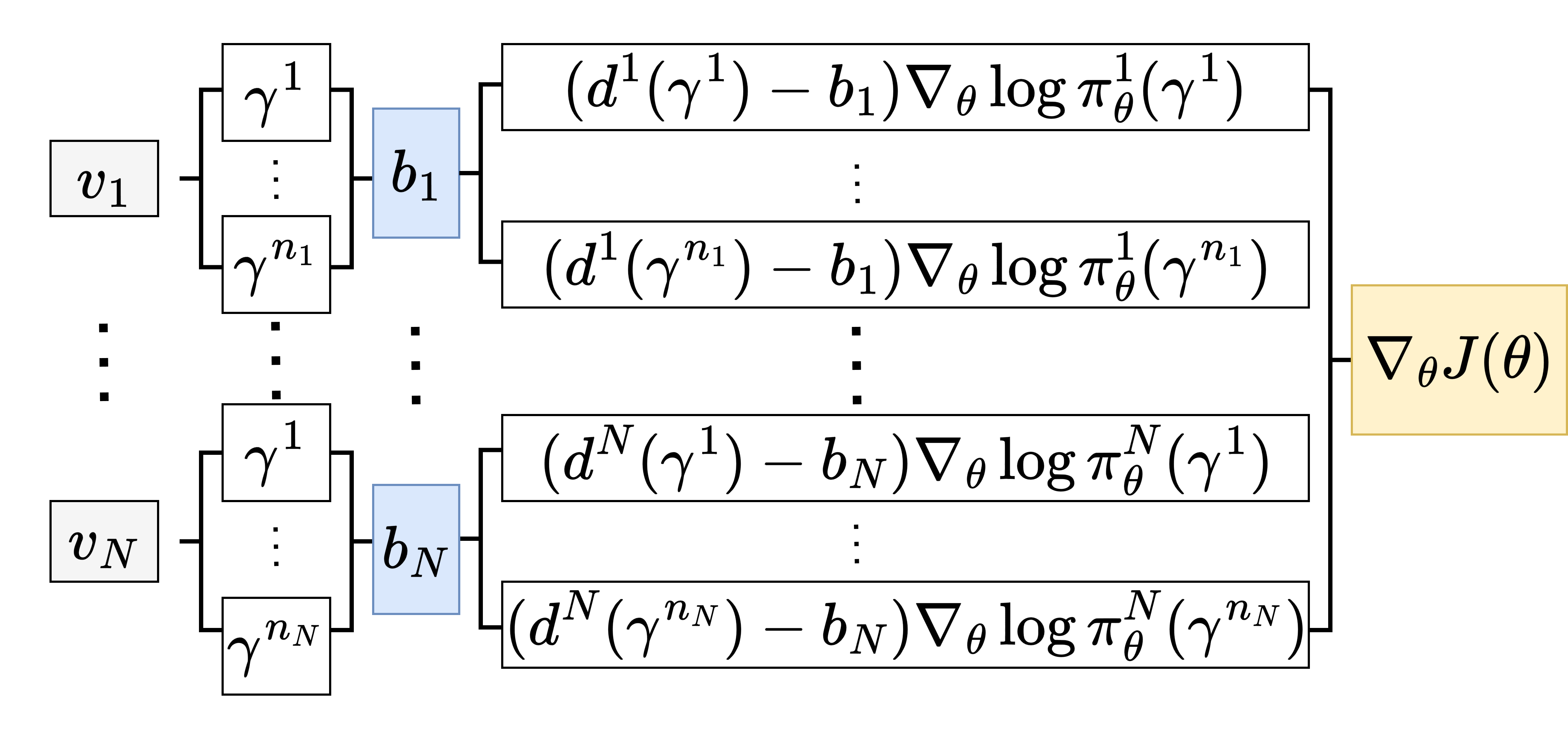}
    \caption{Schematics of unsupervised training of SPN.}
    \label{fig:Unsupervised Learning}
\end{figure}

After supervised pretraining, our deep model is then trained for larger number of nodes $M$ using RL. Specifically, the REINFORCE algorithm \cite{williams1991function} is used to directly train the weights of the model with a baseline inspired by POMO \cite{kwon2020pomo}. Take $J(\theta):= \mathbb{E}_{v_i \sim \mathcal{P}}\left[ \mathbb{E}^\theta_{\gamma \sim \mathcal{G}^i} \left[d^i(\gamma) \right] \right]$. {\color{black} The inner expectation depends on $\theta$ since the paths $\gamma$ are realized via the policy $\pi^i_\theta$}. Using Monte-Carlo approximations for expectations we have,
\begin{align*}
    \nabla_\theta J(\theta) \approx & 
    \frac{1}{Nn_i}\sum_{i=1}^N \sum_{j=1}^{n_i}\big(d^i(\gamma^j)-b_i\big) \nabla_\theta \log \pi^i_\theta(\gamma^j),
\end{align*}
where $\gamma^j \in \mc G^i, 1 \leq j \leq n_i$ are the sampled paths for each instance $v_i$, $b_i = \frac{1}{n_i}\sum_{j'=1}^{n_i}d^i(\gamma^{j'})$ is the baseline which approximates expectation of state-action trajectory costs along the paths $\gamma^j$ and $\log \pi^i_\theta(\gamma^j) = \sum_{k=0}^{M-1}\log \pi_\theta^i \rb{a_{j,k}|s_{j,k}}$. Here the states $s_{j,k}$ represent the nodes $\gamma_k^j$ along the paths $\gamma^j$ and actions $a_{j,k}$ are chosen according to policy $\pi_\theta^i\rb{\cdot|s_{j,k}}$. See Figure \ref{fig:Unsupervised Learning} for a schematics of the unsupervised learning for SPN.

\section{Results}
\begin{figure}[t]
    \centering
    \includegraphics[width=\linewidth]{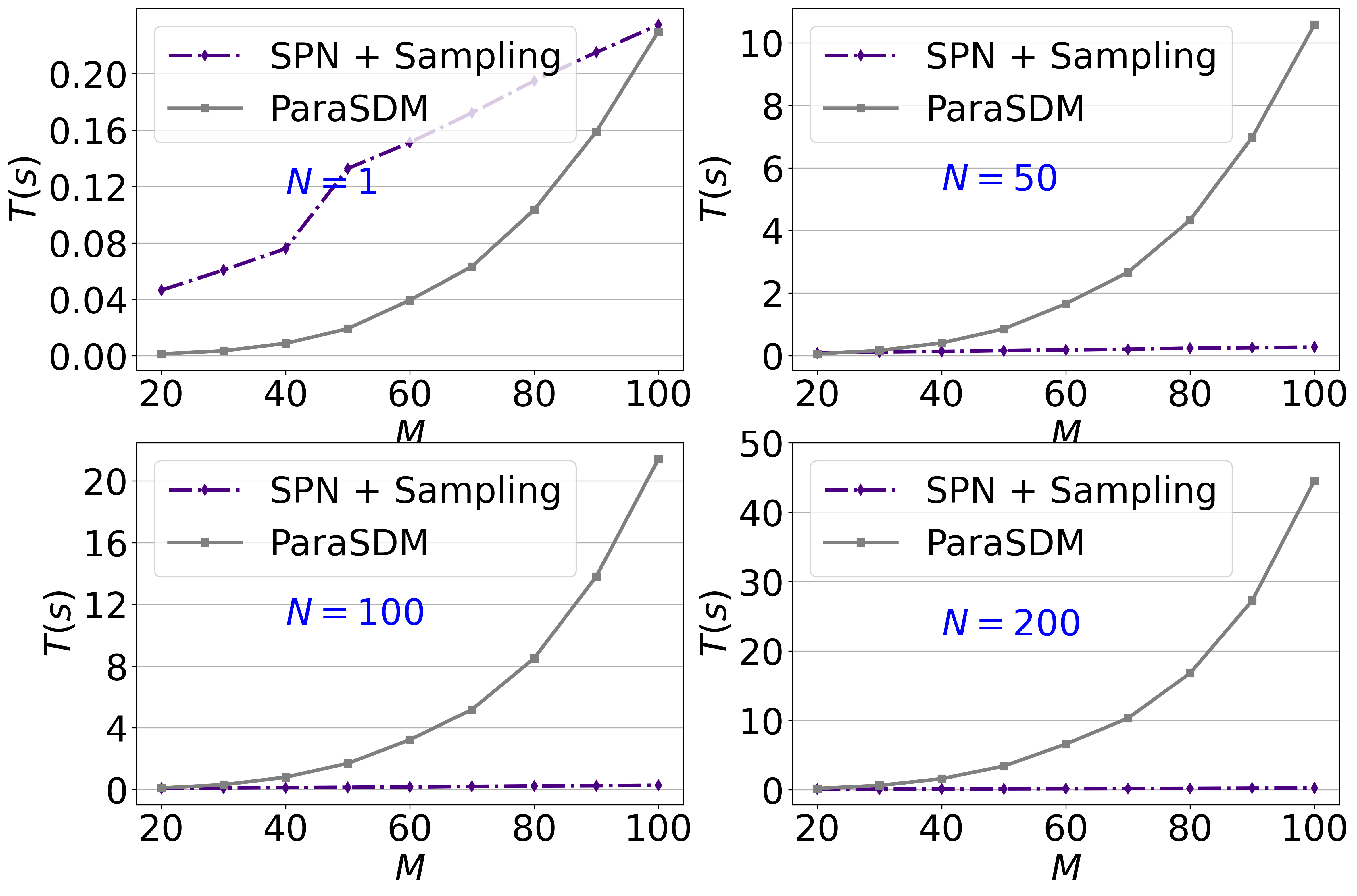}
    \caption{Time Comparison between SPN+Sampling and ParaSDM for Gradient Computation.}
    \label{fig:grad_time_compare}
\end{figure}

\begin{figure}[t]
    \centering
    \includegraphics[width=\linewidth]{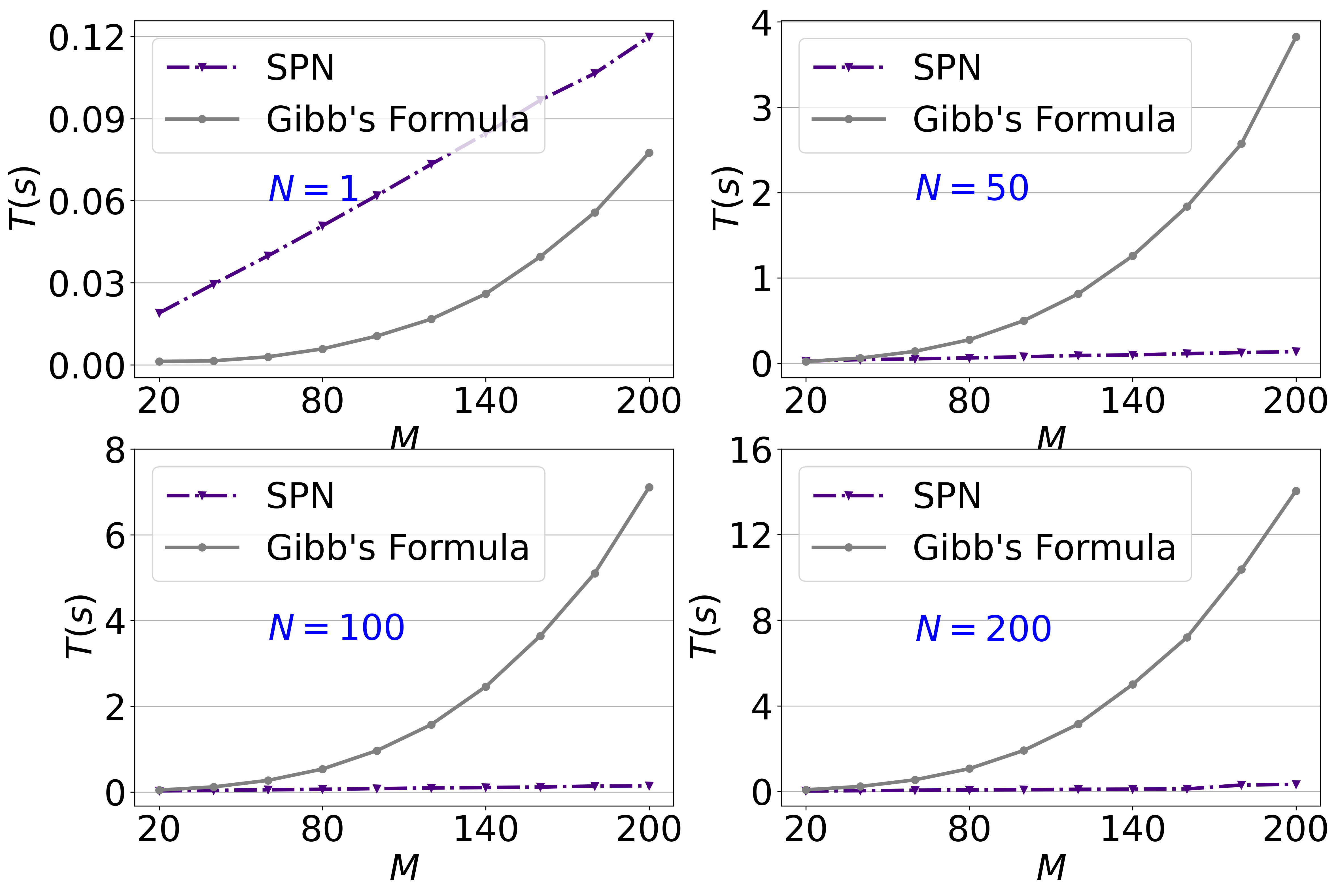}
    \caption{Time Comparison between SPN and ParaSDM for inferring the shortest paths.}
    \label{fig:gibbs_time_compare}
\end{figure}

In this section, we first evaluate the scalability of the proposed SPN framework compared to the original ParaSDM method, focusing on both action policy inference and free energy gradient computation (all the simulations conducted on NVIDIA GH200 120GB GPUs). Figure~\ref{fig:grad_time_compare} demonstrates a substantial speedup in computing gradients of the free energy with respect to parameters using SPN (via top $5$ shortest paths and 10 uniform samples), particularly at larger problem sizes. For instance, at $N = 200, M=100$, SPN achieves $\sim200\times$ speedup relative to ParaSDM. Similarly, Figure~\ref{fig:gibbs_time_compare} compares shortest-path inference times under the stagewise Gibbs policy, showing SPN's greedy inference is approximately $100\times$ faster than ParaSDM for $N = M = 200$. These significant efficiency gains, driven by the architectural innovations in SPN, enable practical optimization of large-scale FLPO instances involving hundreds of agents and nodes—scenarios previously beyond reach due to computational bottlenecks.

\begin{figure*}[t] 

    
    \centering
    
    \begin{subfigure}[b]{0.32\textwidth}
        \includegraphics[width=\textwidth]{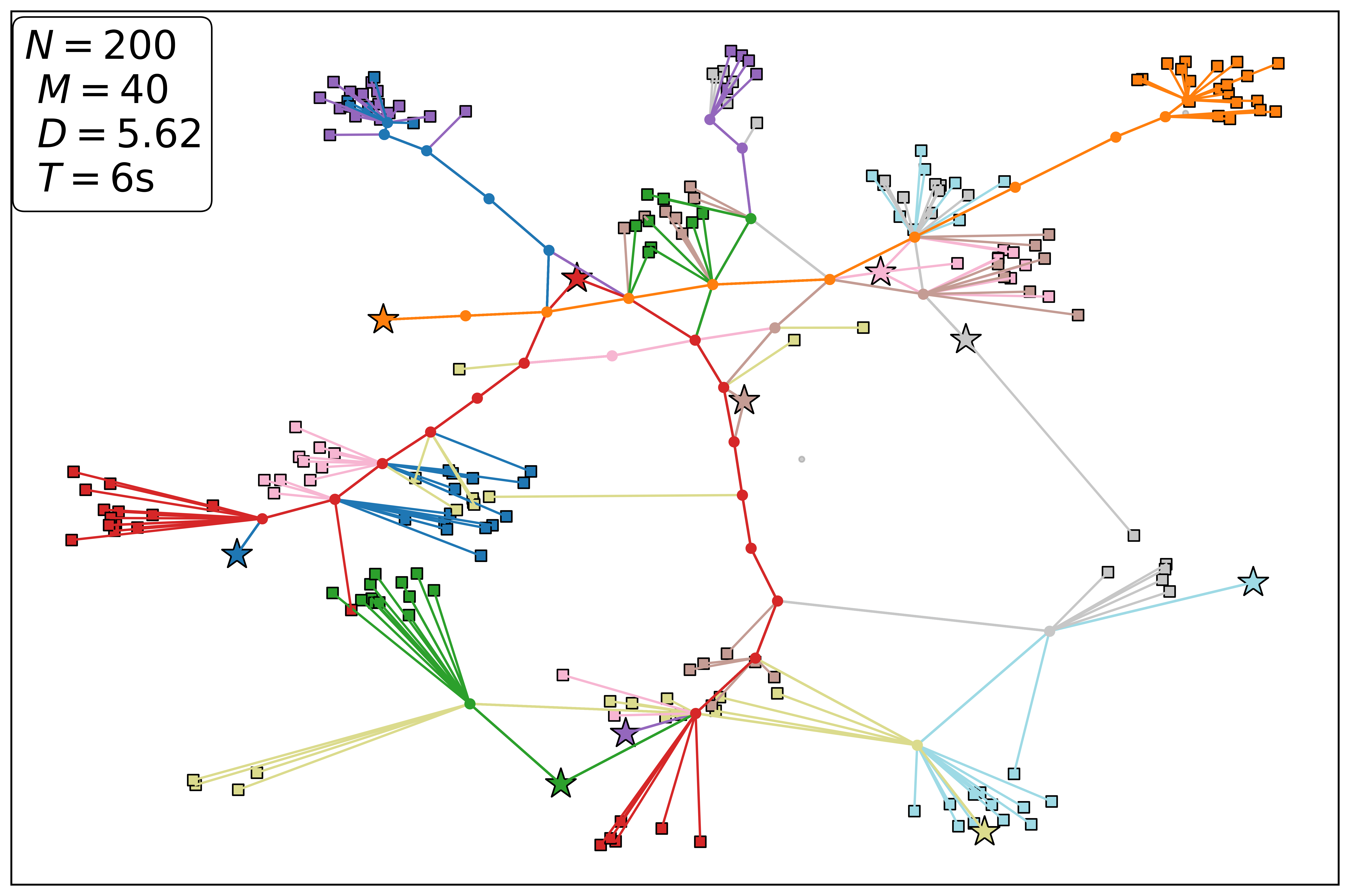}
   
    \end{subfigure}
    \hfill
    \begin{subfigure}[b]{0.32\textwidth}
        \includegraphics[width=\textwidth]{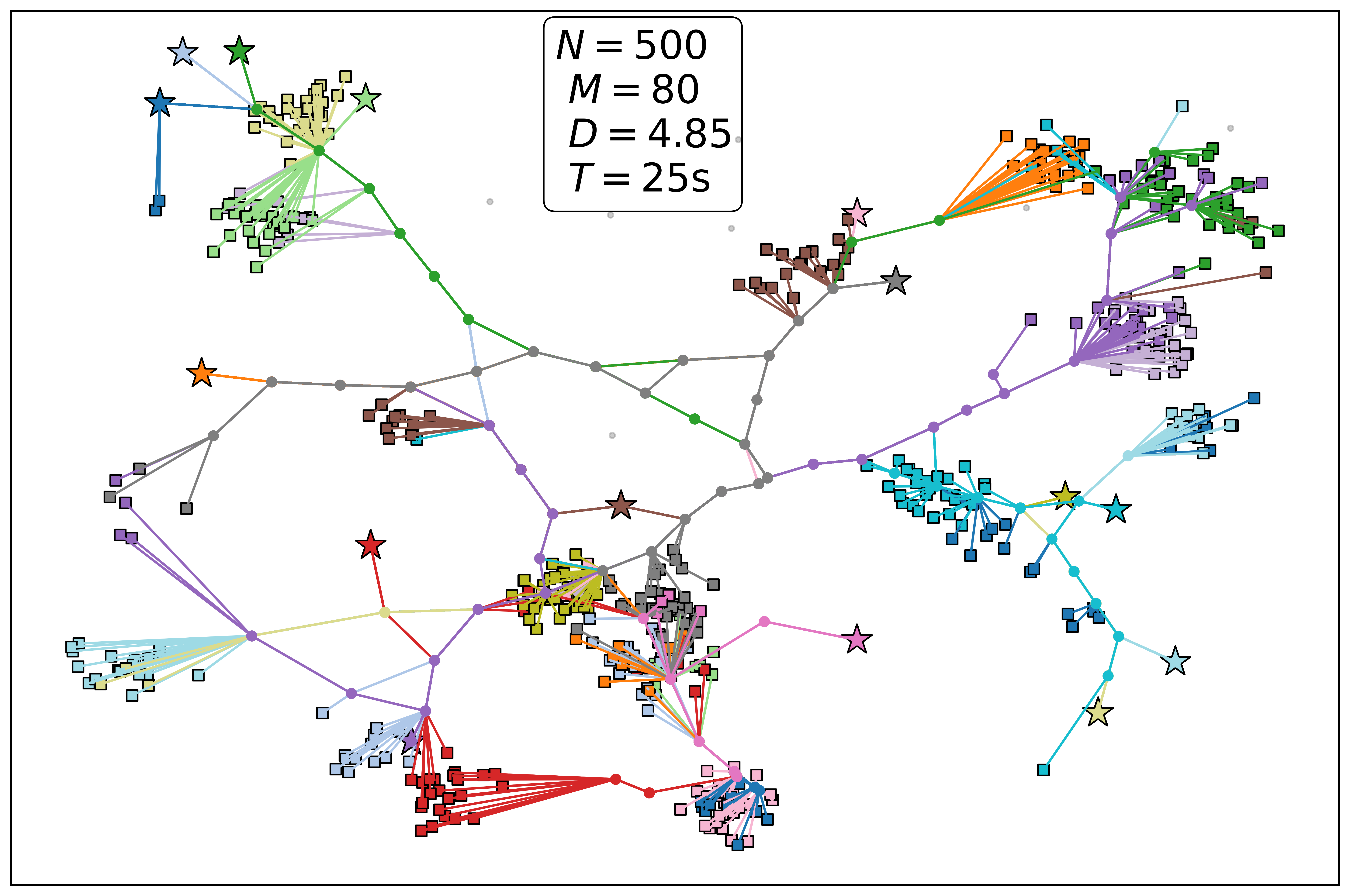}

    \end{subfigure}
    \hfill
    \begin{subfigure}[b]{0.32\textwidth}
        \includegraphics[width=\textwidth]{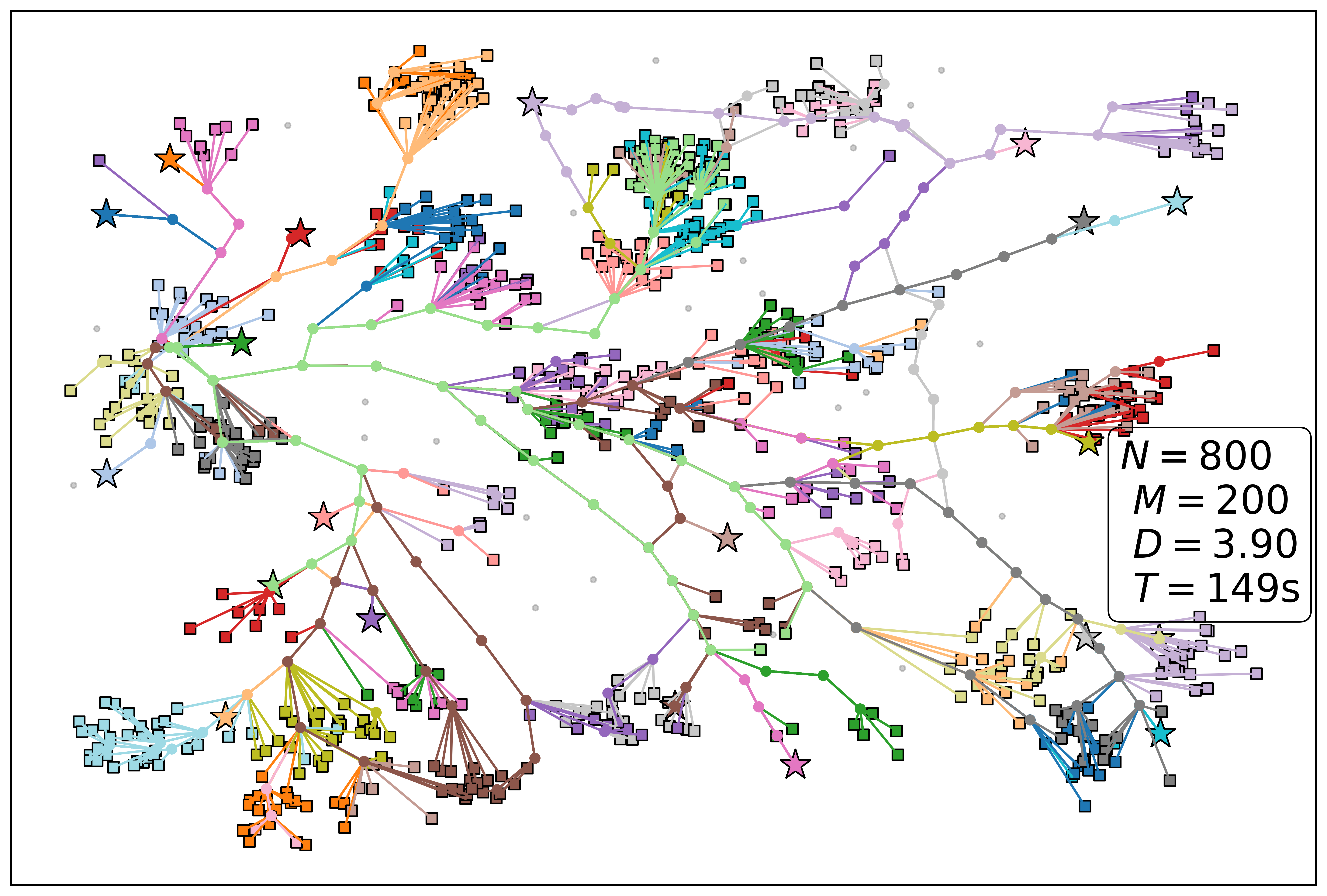}

    \end{subfigure}

    
    
    \begin{subfigure}[b]{0.32\textwidth}
        \includegraphics[width=\textwidth]{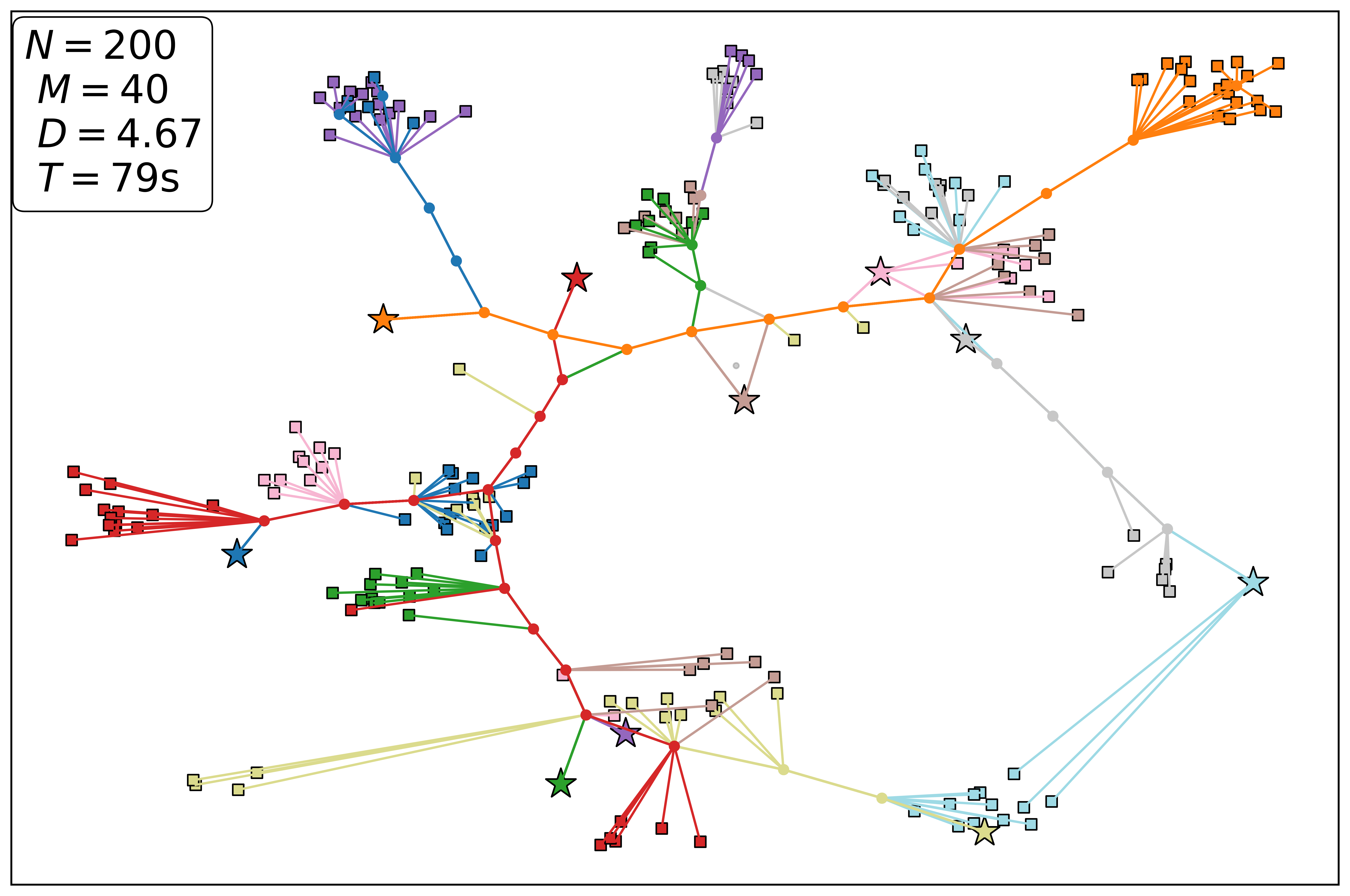}

    \end{subfigure}
    \hfill
    \begin{subfigure}[b]{0.32\textwidth}
        \includegraphics[width=\textwidth]{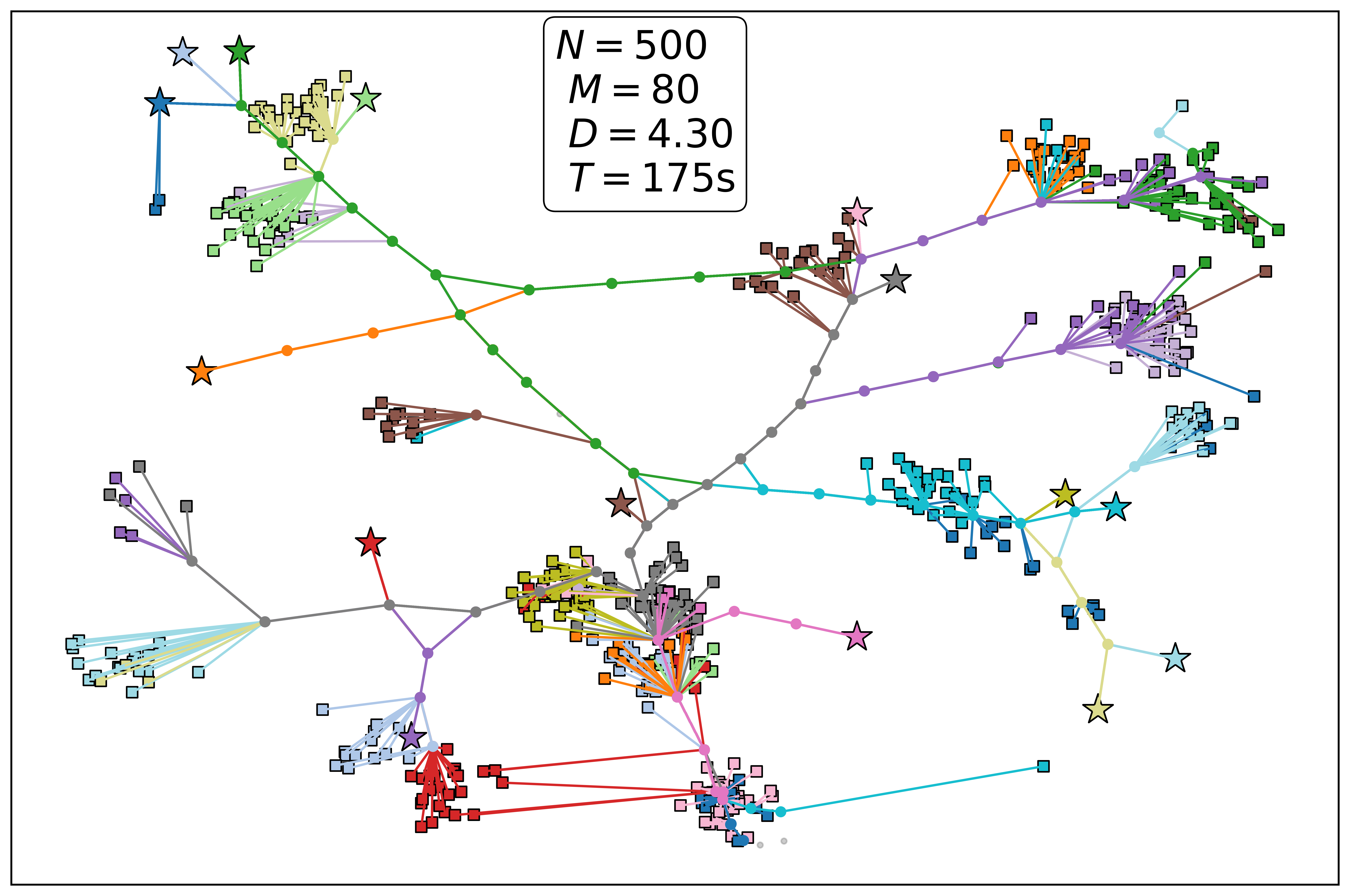}

    \end{subfigure}
    \hfill
    \begin{subfigure}[b]{0.32\textwidth}
        \includegraphics[width=\textwidth]{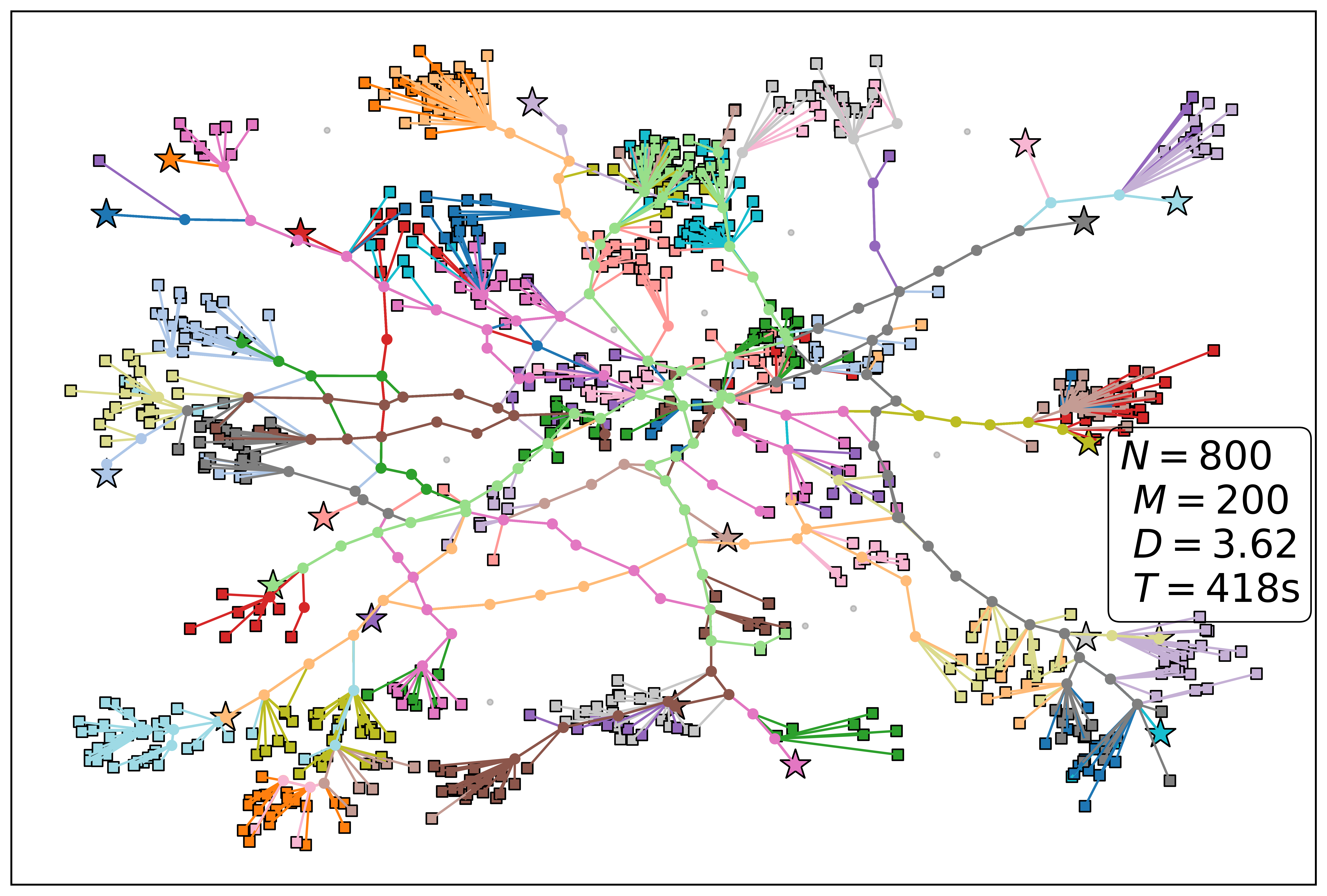}

    \end{subfigure}
    \caption{FLPO Simulation Results (top row: $\beta = 10^3$ bottom row: annealing $\beta$ from $10^{-3}$ to $10^3$ geometrically at rate $10$). \textbf{Squares}: START, \textbf{Stars}: END, \textbf{Dots}: Facility Nodes, \textbf{Colors}: unique for each START-END-Route triplet.
    The cost $D$ in the plot is multiplied by 100 for better readability.}
    \label{fig:FLPO simulations}
\end{figure*}

We showcase our method on three FLPO scenarios in Figure~\ref{fig:FLPO simulations} using both the SPN-only approach (with high $\beta$) and the annealed SPN+Sampling variant. The scenarios are generated within a unit box $\mc D = \rcb{0,1}^2$. Figures illustrate the final locations of facility nodes (charging stations) that agents (drones) must visit along their final routes to reach their goals. Each route is shown in the same color to match its corresponding START-END pair. The SPN-only model achieves up to $10\times$ faster runtime for $N=200$, $M=40$, while the annealed method yields approximately 17\% lower cost $D$. In the larger case of $N=800$, $M=200$, annealing improves cost by 8\% but is $\sim3\times$ slower. This confirms a trade-off: the SPN-only approach is suitable for rapid deployment in large-scale settings, whereas the annealed variant is more robust to initialization and delivers higher-quality solutions. Notably, these problem sizes are beyond the practical limits of ParaSDM baselines, MIP solvers like Gurobi, or standard metaheuristics, which fail to scale accordingly.

We conducted an ablation study to assess the impact of different model and training configurations for SPN. In Phase 1, we compared two decoder architectures—DED and DCAD—under supervised pretraining, with and without annealing applied to the target labels, resulting in four model variants. In Phase 3, we further trained both decoders with and without initializing from Phase 2 to evaluate the contribution of supervised pretraining. The results, detailed in the Appendix \ref{APP:ablation_study}, show that the DED decoder combined with annealed supervision achieves the best performance across a wide range of $M = 10$ to $300$, with an average optimality gap of approximately 6\%—about 4\% better than DCAD with annealing. Supervised pretraining also plays a critical role: for DED, it improves the average optimality gap by 3\%, while the DCAD decoder fails to train effectively using policy gradients alone, yielding extremely poor solutions with 400–500\% gaps. 

We benchmarked Deep FLPO against several widely used heuristic and metaheuristic baselines, including Genetic Algorithm (GA), Simulated Annealing (SA), and the Cross-Entropy Method (CEM), as well as the exact Mixed Integer solver from Gurobi. Due to the computational demands of the baselines, all comparisons were conducted on a moderately small problem instance with $N=10$ agents and $M=4$ spatial nodes. As detailed in Appendix \ref{APP:benchmarks}, Deep FLPO achieved more than a 10$\times$ reduction in cost compared to GA, SA, and CEM, while also running orders of magnitude faster. Compared to Gurobi, our method (at high $\beta$) produced solutions with about $2\%$ higher cost, while with annealing it yields the same cost as Gurobi approximately 1500$\times$ faster. These results highlight the efficiency and scalability of Deep FLPO in solving ParaSDM problems and establish a new state of the art for this class of optimization tasks.

\section*{Acknowledgments}
This work was supported by NASA under Grant 80NSSC22M0070. Computational resources were provided by the Advanced Cyberinfrastructure Coordination Ecosystem: Services \& Support (ACCESS) program (allocation CIS250450) on Delta AI at the National Center for Supercomputing Applications (NCSA), with support from the National Science Foundation (NSF).

\bibliography{aaai2026}

\onecolumn
\section{Appendix}
\subsection{Ablation Results}
\label{APP:ablation_study}
In the ablation study, we compared the effect of choosing different decoder architectures for SPN as well as the effect of annealing in supervised training and the effect of pretraining for the RL training phase. Table \ref{tab:ablation} shows how different approaches compare to the true cost when inferred in greedy or beam search (beam width = 5). The experiments is done across varying network sizes, with $N=128$ and $M=10 \to 300$ to assess the generalizability of each model. DED and DCAD refer to the two different decoder architectures proposed in section \ref{Sec: SPN}. 

\begin{table*}[tbhp]
\centering
\begin{tabular}{llccccc}
\toprule
Model & $M$ & 10 & 50 & 100 & 200 & 300 \\
\midrule
\textbf{True Cost}     & \textbf{—}      & \textbf{0.191} & \textbf{0.083} & \textbf{0.063} & \textbf{0.042} & \textbf{0.035} \\
\midrule
\multirow{2}{*}{DED, Anneal} 
  & BS     & 0.191 & 0.084 & 0.064 & 0.046 & 0.041 \\
  & Greedy & 0.207 & 0.085 & 0.066 & 0.049 & 0.044 \\
\midrule
\multirow{2}{*}{DED, No Anneal} 
  & BS     & 0.194 & 0.084 & 0.064 & 0.046 & 0.041 \\
  & Greedy & 0.214 & 0.085 & 0.065 & 0.048 & 0.043 \\
\midrule
\multirow{2}{*}{DED, No Pre-train} 
  & BS     & 0.193 & 0.084 & 0.065 & 0.047 & 0.043 \\
  & Greedy & 0.251 & 0.089 & 0.068 & 0.050 & 0.046 \\
\midrule
\multirow{2}{*}{DCAD, Anneal} 
  & BS     & 0.236 & 0.084 & 0.064 & 0.045 & 0.040 \\
  & Greedy & 0.288 & 0.086 & 0.065 & 0.047 & 0.042 \\
\midrule
\multirow{2}{*}{DCAD, No Anneal} 
  & BS     & 0.241 & 0.084 & 0.064 & 0.045 & 0.040 \\
  & Greedy & 0.311 & 0.086 & 0.066 & 0.047 & 0.042 \\
\midrule
\multirow{2}{*}{DCAD, No Pre-train} 
  & BS     & 0.335 & 0.334 & 0.338 & 0.302 & 0.343 \\
  & Greedy & 0.335 & 0.334 & 0.338 & 0.302 & 0.343 \\
\bottomrule
\end{tabular}
\caption{Beam Search (BS) and Greedy Costs for Different Models and $M$ Values.}
\label{tab:ablation}
\end{table*}

\begin{figure}[tbhp]
    \centering
    \includegraphics[width=0.3\linewidth]{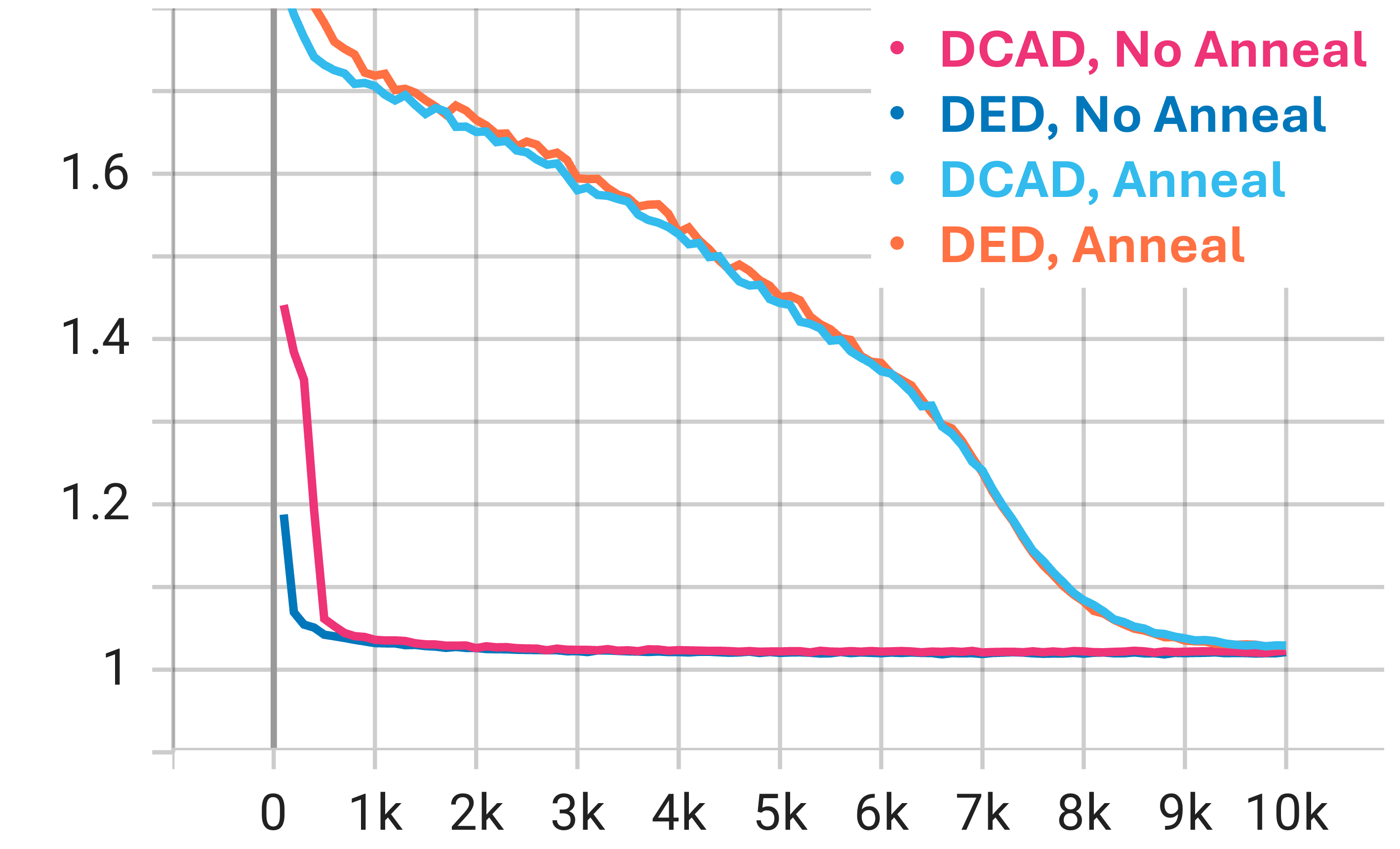}
    \includegraphics[width=0.3\linewidth]{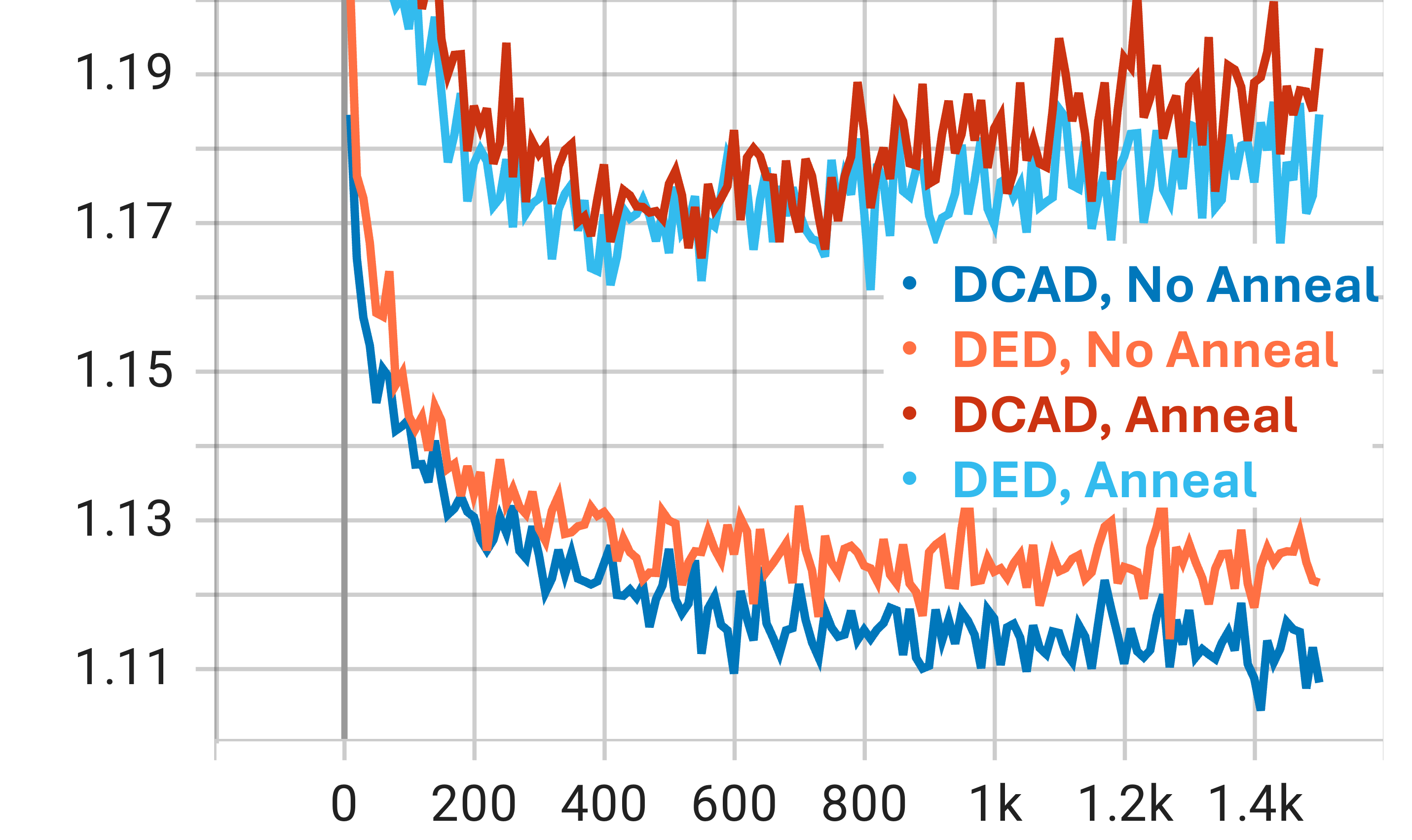}
    \caption{Mean Route Length vs. Epochs - top: Phase 1, bottom: Phase 2.}
    \label{fig:supervised training curves}
\end{figure}

\begin{figure}[tbhp]
    \centering
    \includegraphics[width=0.3\linewidth]{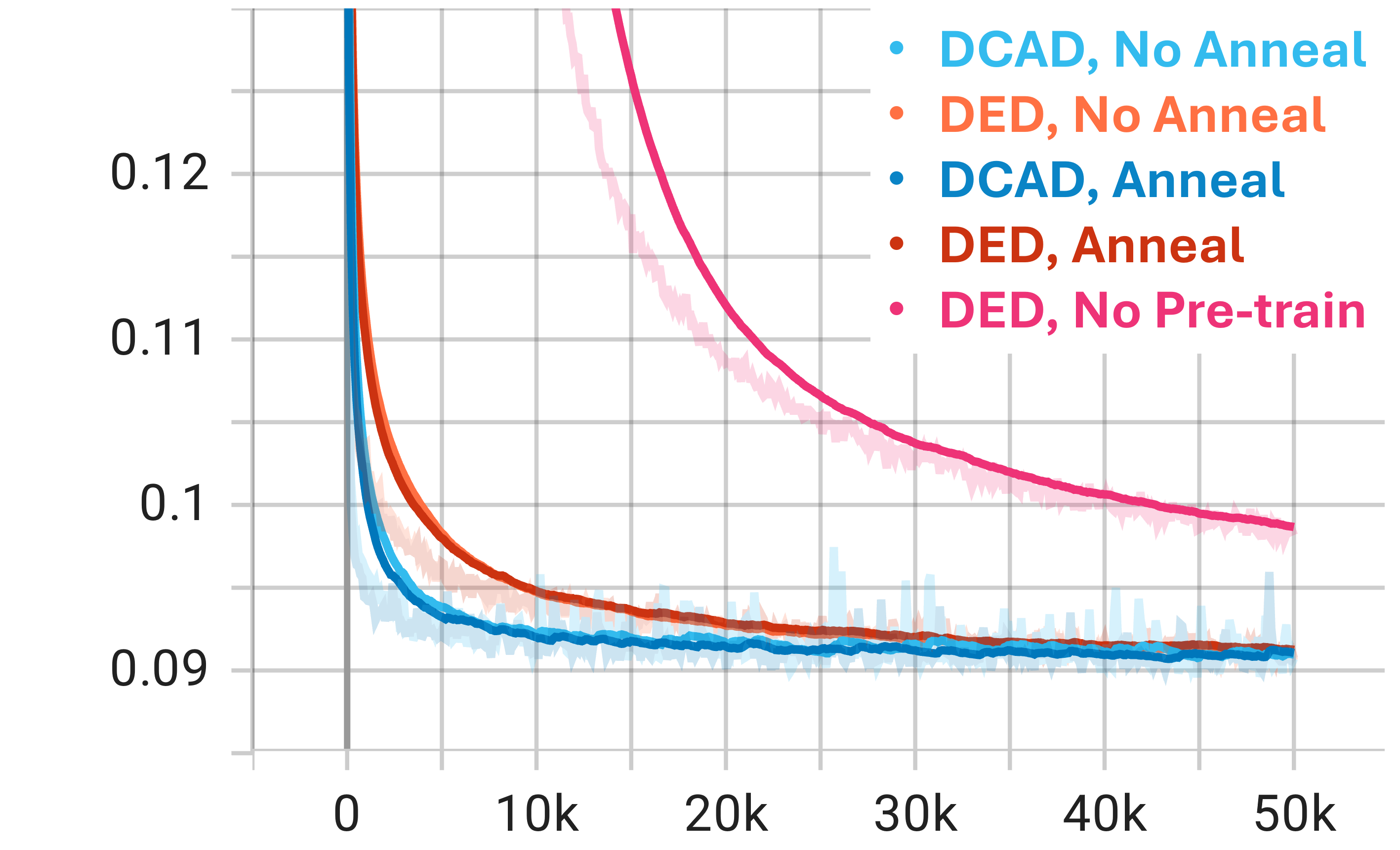}
    \includegraphics[width=0.3\linewidth]{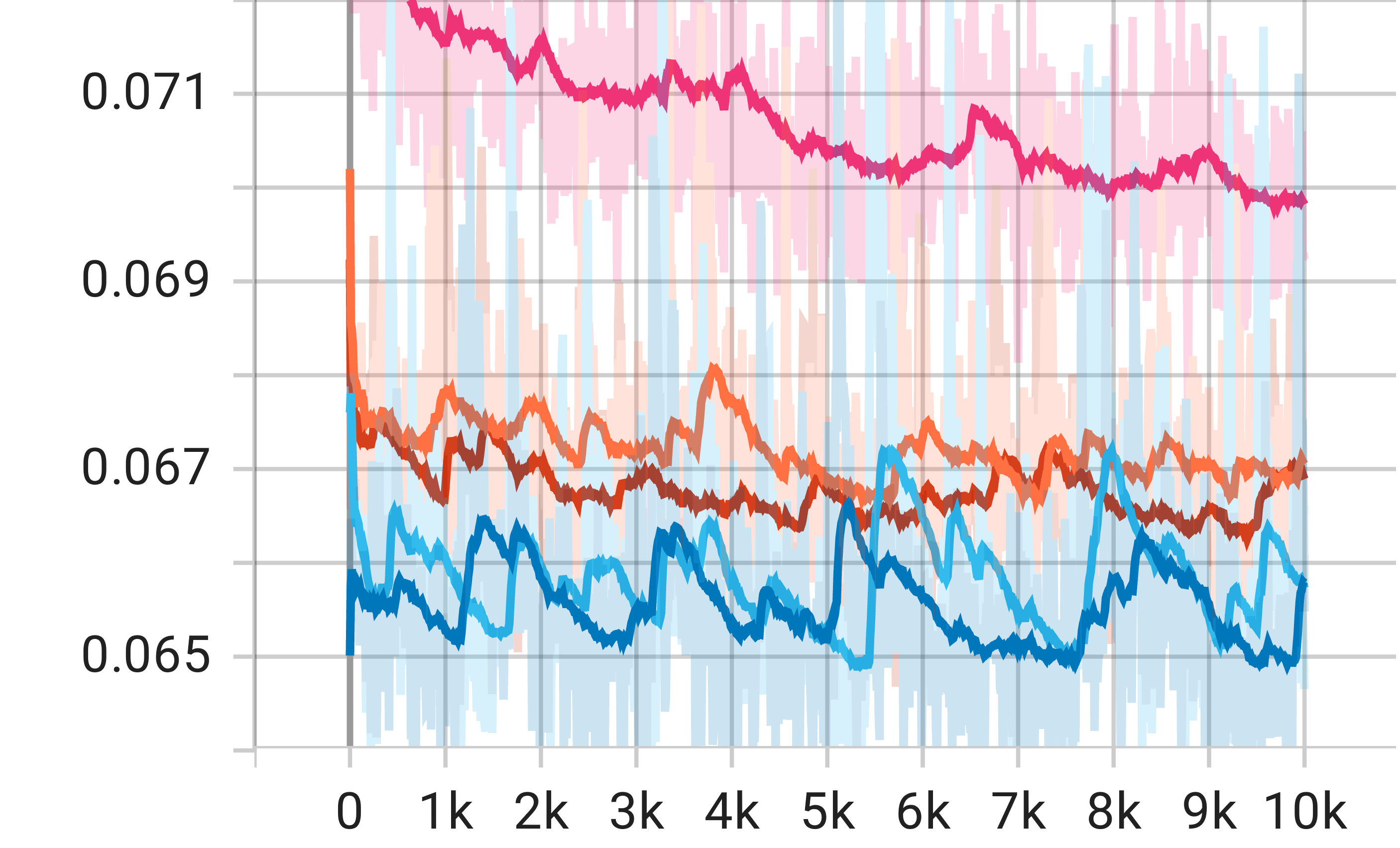}
    \caption{Smoothed Mean Route Length vs. Epochs - top: Phase 3, bottom: Phase 4 (same legend).}
    \label{fig:RL training curves}
\end{figure}

\begin{figure*}[tbhp]
    \centering
    \begin{subfigure}[b]{0.48\textwidth}
        \includegraphics[width=\textwidth]{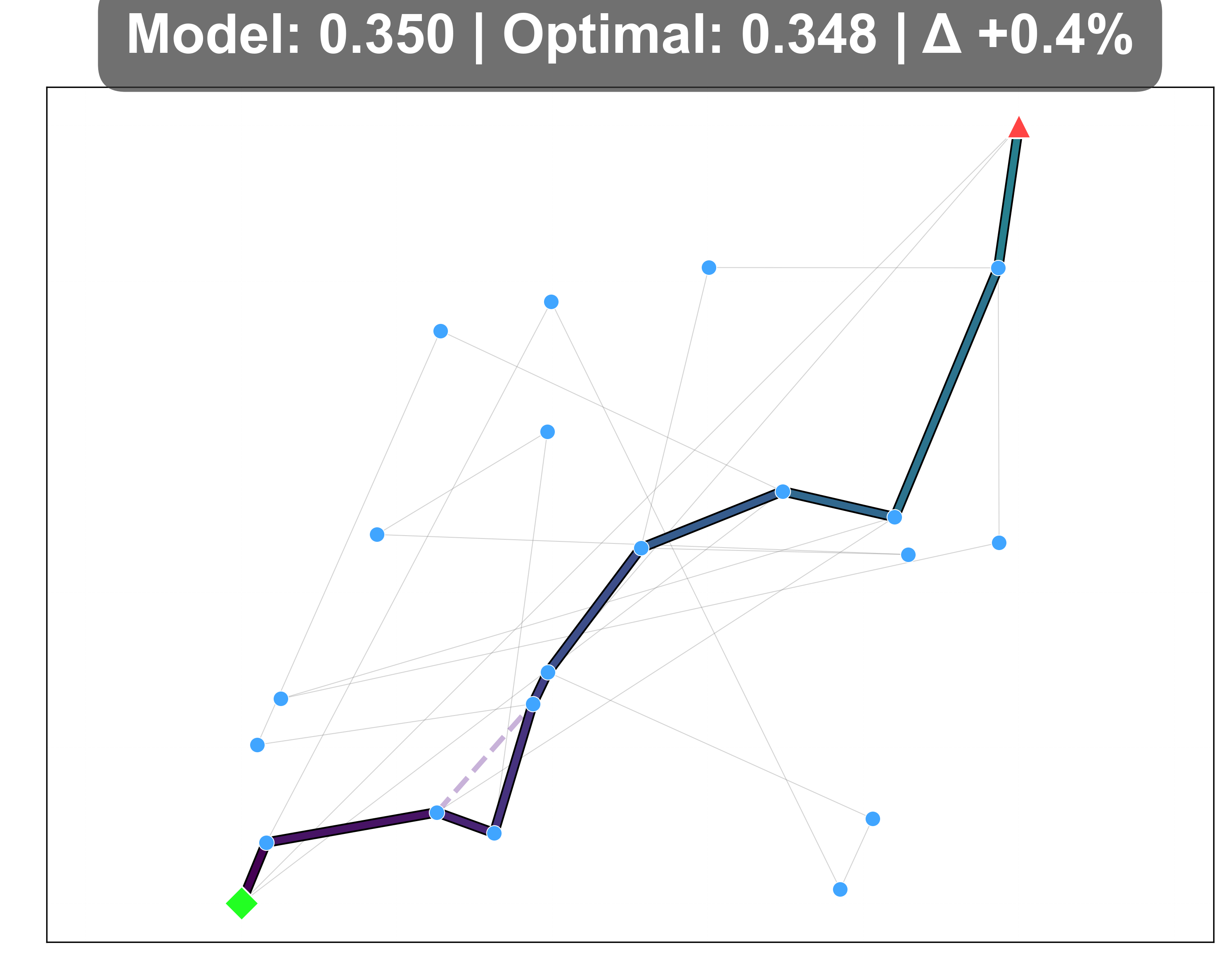}
        
    \end{subfigure}
    \hfill
    \begin{subfigure}[b]{0.48\textwidth}
        \includegraphics[width=\textwidth]{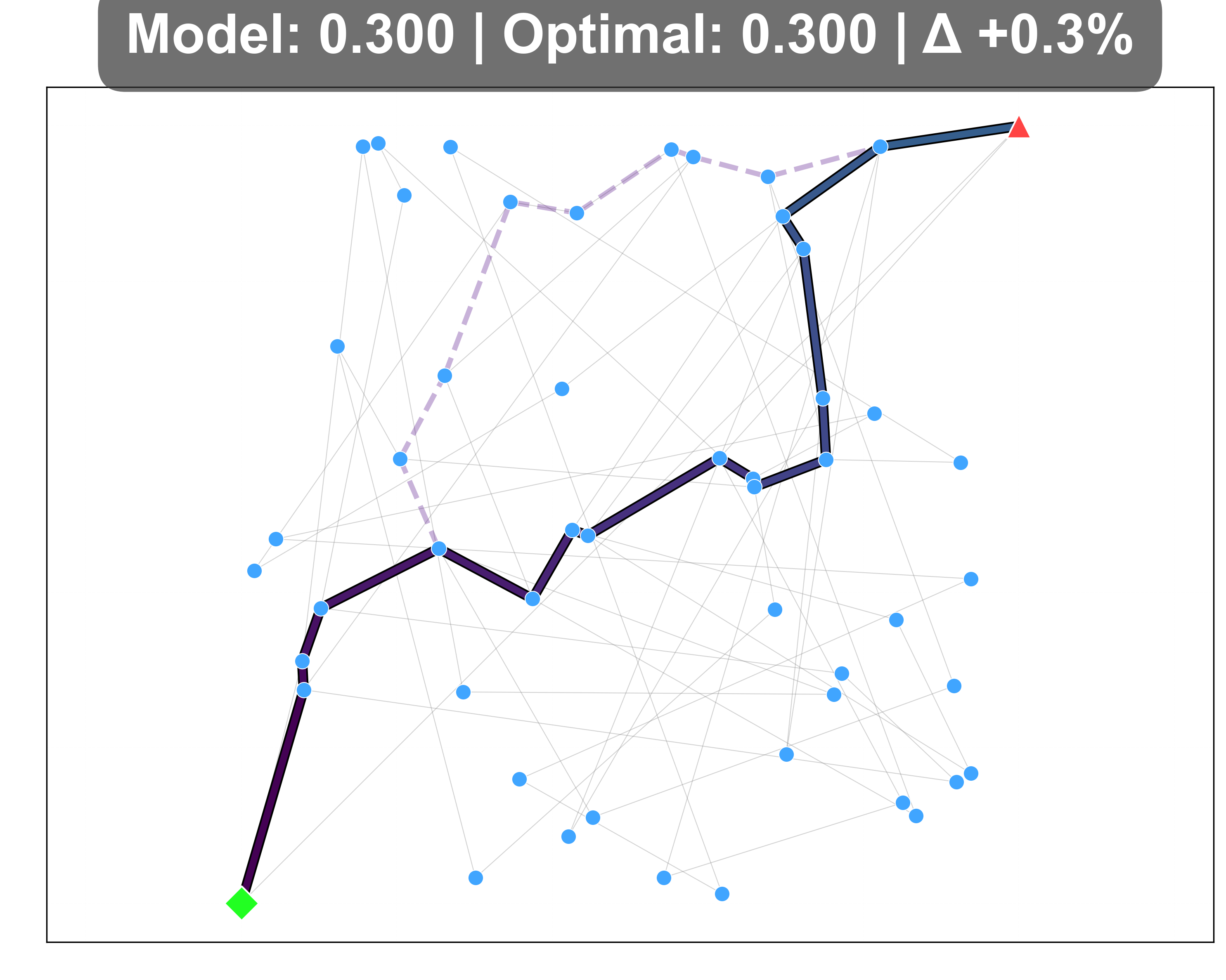}
       
    \end{subfigure}
    
    \begin{subfigure}[b]{0.48\textwidth}
        \includegraphics[width=\textwidth]{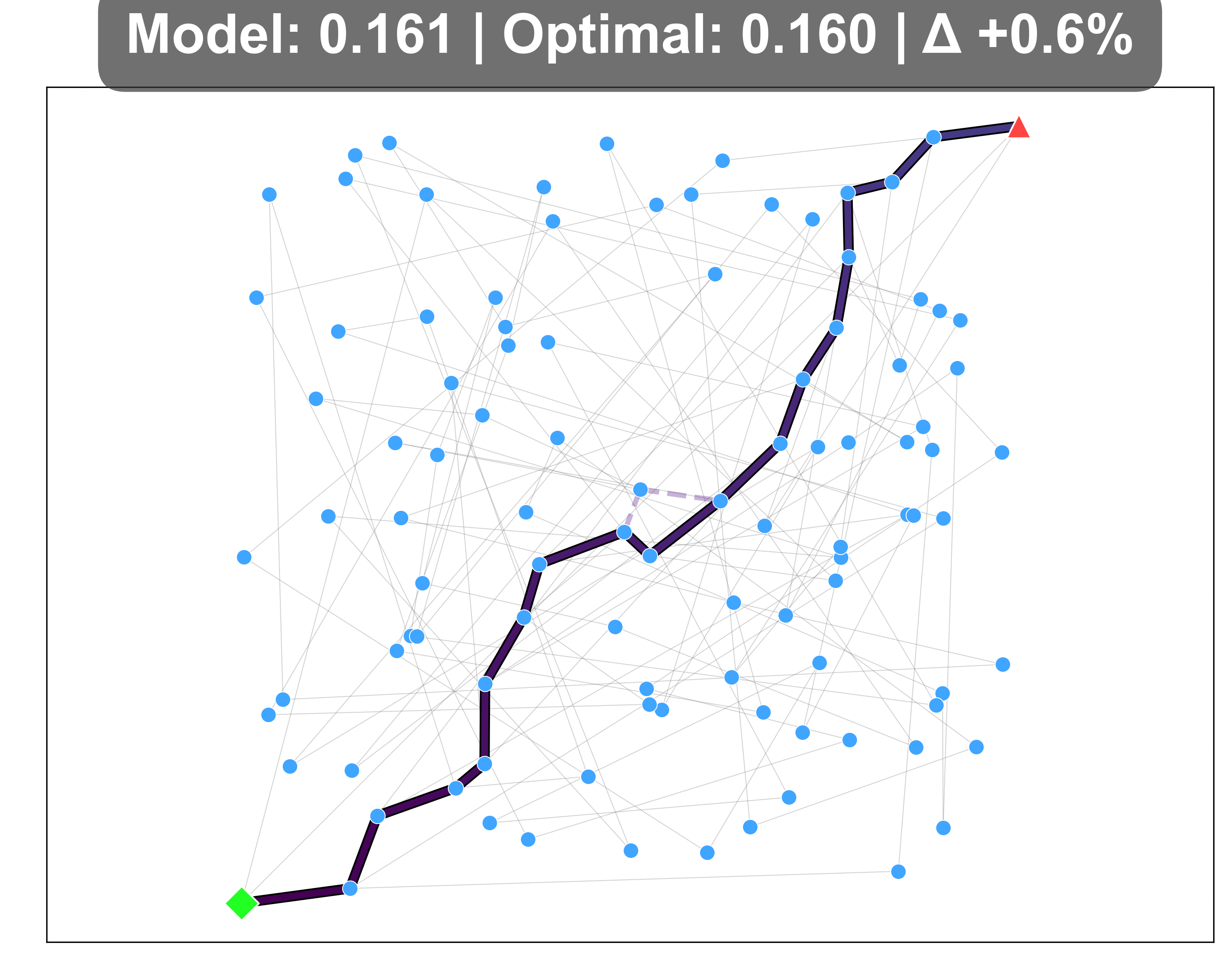}
       
    \end{subfigure}
    \hfill
    \begin{subfigure}[b]{0.48\textwidth}
        \includegraphics[width=\textwidth]{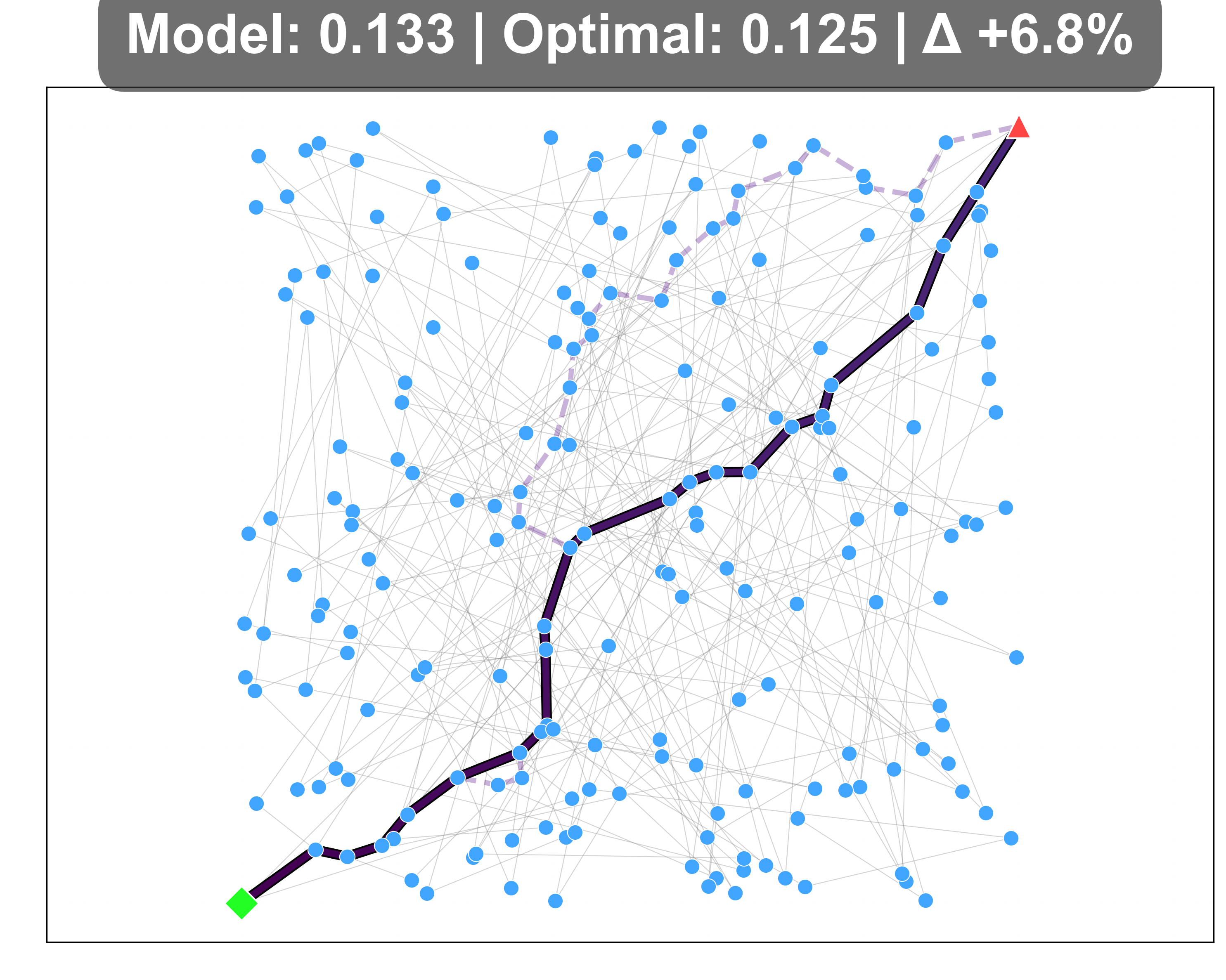}
    \end{subfigure}
    
    \caption{SPN Performance on $M=20, 50, 100, 200$ sized networks, compared with true shortest paths (the dashed line). While the graph is fully connected, only a portion of graph edges have been shown for better visibility.}
    \label{Fig: SPN Examples}
\end{figure*}

We can also see the plots of Mean Route Length vs. epochs for supervised training (Phase 1 and 2) in Figure \ref{fig:supervised training curves}. The same plots in the RL phase of training can be seen in Figure \ref{fig:RL training curves}. Also, Figure \ref{Fig: SPN Examples} shows the SPN performance across various problem instances as an example.

\subsection{State Value and State-Action Value Functions}
\label{APP:value_funcs_free_energy}

We consider the problem setup in Section~\ref{sec:problem_formulation} and minimization of $F_\beta$ in Eq~\eqref{eq:free_energy}. For each agent, Markov property implies that path associations $p^i\rb{\gamma}$ can be written as $p^i\rb{\gamma} = \prod_{k=0}^{M-1} p^i\rb{\gamma_{k+1}|\gamma_k}, \ \gamma_0 = s_i$ with $p^i\rb{\Delta_i|\gamma_M} = 1, \forall \gamma_M \in \Gamma_M^i$. Further, suppose the path costs $d^i\rb{\gamma}$ can be dissociated as $d^i\rb{\gamma} = \sum_{k=0}^{M} d^i\rb{\gamma_k, \gamma_{k+1}}$. For each $i$ and $1\leq k \leq M$, we define the
parameterized state value function $V_k^i$ and parametrized state-action value function $\Lambda_k^i$ as below:
\begin{align}
    V^i\rb{\gamma_k} & := \sum_{\gamma_{k+1},\dots,\gamma_M} \prod_{t=k}^{M} p^i(\gamma_{t+1} | \gamma_t) \cdot \sum_{t=k}^M \rcb{d^i\rb{\gamma_t,\gamma_{t+1}} + \frac{1}{\beta} \log p^i\rb{\gamma_{t+1} | \gamma_t}},    \label{eq:value_def} \\
    \Lambda^i\rb{\gamma_k, \gamma_{k+1}} & := d^i\rb{\gamma_k, \gamma_{k+1}} + V^i\rb{\gamma_{k+1}}. \label{eq:state_action_def}
\end{align}
Using the principle of optimality, the first order necessary condition on state-value function $\frac{\partial V^i_k\rb{\gamma_k}}{\partial p^i_k\rb{\gamma_k|\gamma_{k+1}}} = 0$ at each $\beta$ yields the following recursive form,
\begin{align}
    V^i\rb{\gamma_k} & = -\frac{1}{\beta} \log {\sum_{\gamma_{k+1}'} e^{-\beta \Lambda^i\rb{\gamma_k, \gamma_{k+1}'}}} \forall k, \forall i \label{eq:value_recursive}
\end{align}
where $\Lambda_k^i$ satisfies Eq~\eqref{eq:state_action_def}. The expression for $F_\beta$ is shown to be $F_\beta = \sum_{i=1}^N \rho_i V^i\rb{s_i}.$

\subsection{Gradient of $\nabla_{\mc Y} F_\beta$ in terms of $V^i\rb{\gamma_k}$}
\label{APP:gradF_DP}
Using shorthand notations, $V_k^i = V\rb{\gamma_k}, d_k^i = d^i\rb{\gamma_k, \gamma_{k+1}}, p_k^i = p^i\rb{\gamma_{k+1}|\gamma_k}$, then for each instance $\mc Y$, and $\forall i, \forall k,$ the gradient is obtained in terms of state value function $V_k^i\rb{\gamma_k}$, as below: 
\begin{align}
    \nabla_{\mc Y}F_\beta  & = \sum_{i=1}^N \rho_i \nabla_{\mc Y}V_0^i = 0, \forall j, \label{eq:grad_Free_energy} \\
    \nabla_{\mc Y} V_{k}^i & = \sum_{\gamma_{k+1}} p^i\rb{\gamma_{k+1}|\gamma_k} \cdot 
    \nabla_{\mc Y}\rcb{d_k^i + V_{k+1}^i}, \label{eq:grad_value}\\ 
    p^i_k  & = \frac{\exp\rcb{-\beta {\Lambda^i\rb{\gamma_k, \gamma_{k+1}}}}}{\sum_{\gamma'_{k+1}}\exp\rcb{-\beta {\Lambda^i\rb{\gamma_k, \gamma'_{k+1}}}}}. \label{eq:gibbs_stagewise}
\end{align}

\subsection{Worst case computational complexity of $\nabla_{\mc Y}F_\beta$ via Eq~\eqref{eq:gradF_y_1} is $O\rb{NM^2\sum_{k=1}^M \binom{M}{k} k!}$}
\label{APP:complexity_grad_direct}
\begin{enumerate}
    \item Each path gradient $\nabla_{\mc Y}{d^i\rb{\gamma}}$, $\forall \gamma$, requires $O\rb{M^2}$ operations which must be repeated for a total of $O\rb{\sum_{k=1}^M \binom{M}{k}k!}$ paths.
    \item The Gibbs distribution $p^i\rb{\gamma}, \forall \gamma$ incurs the cost of $O\rb{M\sum_{k=1}^M \binom{M}{k}k!}$. 
    \item Computing $\nabla_{\mc Y} d^i\rb{\gamma} p^i\rb{\gamma}$ and summing all the terms requires another $O\rb{\sum_{k=1}^M \binom{M}{k} k!}$ operations. This results in a total of $O\rb{M^2\sum_{k=1}^M \binom{M}{k}k!}$ operations.
    \item Further, these operations must be performed for each agent $1 \leq i \leq N$.
\end{enumerate}

\subsection{Worst case computational complexity of $\nabla_{\mc Y} F_\beta$ via \eqref{eq:grad_value}, \eqref{eq:gibbs_stagewise} is $O(NM^4)$}
\label{APP:complexity_grad_stagewise}
\begin{enumerate}
    \item Obtaining the Gibbs distribution \( p_k^i(\gamma_{k+1} \mid \gamma_k) \) requires \( O(M^2) \) operations. Multiplication with \( \nabla_{y_j}\rcb{d_k^i + V_{k+1}^i} \) adds another \( O(M^2) \) operations for each \( j \). The summation \( \sum_{\gamma_{k+1}} \) requires an additional \( M \) operations effectively requiring $O\rb{M^2}$ operations for obtaining $\nabla_{y_j} V^i_k$. 
    \item The above operations must be repeated for each $1\leq j \leq M$, resulting in \( O(M^3) \) operations for each \( k \). Further, repeating these operations for each \( 1 \leq k \leq M \) results in a total of \( O(M^4) \) operations.
    \item The above operations must be performed for each agent $1 \leq i \leq N$.
\end{enumerate}

\subsection{Benchmarks}
\label{APP:benchmarks}

We compare two variants of our approach: FLPO with only the SPN component, which is solved at a high value of \( \beta \), and FLPO with SPN and sampling, where \( \beta \) is annealed from \( 10^{-3} \) to \( 10^4 \) at a geometric rate of 10. These approaches are evaluated against baseline methods including Genetic Algorithm (GA), Simulated Annealing (SA), the Cross Entropy Method (CEM), and the exact Mixed Integer Programming (MIP) solver from Gurobi. These benchmarks were implemented on a system with NVIDIA RTX 4070 GPU and 13th Generation Intel Core i7 CPU.

All methods are tested on a problem instance with \( N = 10 \) agent start locations, 2 end locations, and \( M = 4 \) intermediate nodes. Each algorithm is run 10 times, and we report the \textit{minimum cost and runtime} across runs for each method—except for Gurobi, where only cost is reported due to its exceptionally high runtime compared to the other solvers. Key hyperparameters (shown in the table) were tuned to yield the best performance for each method.

We observe that both of our approaches achieve costs that are 10$\times$ lower than those of GA, SA, and CEM, while also running orders of magnitude faster. The SPN-only variant produces a cost within 2\% of the Gurobi baseline, and the annealing variant achieves similar cost. However, Gurobi incurs significantly higher computational overhead compared to our methods.

\begin{table}[tbhp]
\centering
\small  
\begin{tabular}{@{}lllll@{}} 
\hline
\textbf{Method} & \textbf{Key Parameters} & \textbf{Values} & \textbf{Cost} & \textbf{Runtime}  \\ \hline
GA  & Population Size & 100 & & \\
    & Generations & 8000 & $0.724$ & $\sim 3$m\\
    & Crossover & 0.5 & & \\
    & Mutation Rate & 0.3 & & \\ \hline
SA  & Initial Temperature & 1.0 & &  \\
    & Cooling Rate & 0.995 & $1.064$ & $\sim 10$s\\
    & Number of Iterations & 50000 & & \\ 
    \hline
CEM & Population Size  & 100 & & \\
    & Number of Iterations & 2000 & $0.931$ & $\sim 2.5$s \\
    & Elite Fraction & 0.2 & & \\ \hline
Deep FLPO & $\beta$ & $10^4$ & & \\
(SPN-only at high beta)    & Number of Uniform Samples & 8 & &  \\
    & Number of Iterations & 100 & $0.063$ & $\sim 1$s \\
    & Stepsize & 0.01 & & \\
    & Convergence tolerance & 0.001 & & \\ \hline
Deep FLPO & $\beta$ & $10^{-3} \rightarrow 10^4$ & & \\
(SPN+Sampling with annealing)      & Number of Uniform Samples & 8 & &  \\
    & Number of Iterations & 100 & $0.062$ & $\sim 20$s \\
    & Stepsize & 0.01 & & \\
    & Convergence tolerance & 0.001 & & \\ \hline
\textbf{Gurobi} & Gap Tolerance & $\mathbf{1\%}$ & $\mathbf{0.062}$ & $\sim 53$m \\ \hline
\end{tabular}
\caption{Benchmark comparison of our method against various heuristic and metaheuristics.}
\label{tab:benchmark_hyperparams}
\end{table}

\end{document}